\documentclass[preprint,11pt,authoryear]{elsarticle}
\usepackage{lineno}
\usepackage{bm}
\usepackage[ruled,linesnumbered]{algorithm2e}
\usepackage{booktabs} 
\usepackage{siunitx} 
\usepackage[table]{xcolor}
\usepackage{color}
\usepackage[left=2.5cm,right=2.5cm,top=3cm,bottom=3cm]{geometry}
\usepackage{svg}
\usepackage{amsmath}

\usepackage{multirow}

\usepackage{booktabs}
\usepackage{graphicx}
\usepackage{subfigure}
\usepackage{epstopdf}
\usepackage{picinpar}
\usepackage{amsmath}
\usepackage{amsfonts}
\usepackage[mathscr]{euscript}
\usepackage{calligra}
\usepackage{array}
\usepackage[authoryear]{natbib} 
\usepackage{setspace}
\usepackage{setspace}
\usepackage{lipsum} 
\usepackage{adjustbox}

\usepackage{tabularx}

\makeatletter
\def\ps@pprintTitle{} 
\makeatother

\usepackage{titlesec}
\titlespacing{\section}{0pt}{6pt}{6pt}
\titlespacing{\subsection}{0pt}{3pt}{3pt}
\titlespacing{\subsubsection}{0pt}{2pt}{2pt}

\DeclareMathAlphabet{\mathpzc}{OT1}{pzc}{m}{it}

\newdefinition{assumption}{Assumption}

\newdefinition{definition}{Definition}
\newdefinition{remark}{Remark}
\newproof{proof}{Proof}

\biboptions{authoryear,compress}

\let\OLDthebibliography\thebibliography
\renewcommand\thebibliography[1]{
  \OLDthebibliography{#1}
  \setlength{\parskip}{0pt}
  \setlength{\itemsep}{0pt plus 0.3ex}
}
\usepackage[colorlinks=true,
            linkcolor=blue,
            citecolor=blue,
            urlcolor=blue]{hyperref}  

\usepackage{cleveref}

\begin{document}

\begin{frontmatter}

\title{CogDrive: Cognition-Driven Multimodal Prediction-Planning Fusion \\ for Safe Autonomy}

\author[a,b]{Heye Huang\textsuperscript{\dag}}
\ead{heyeh@mit.edu}

\author[c]{Yibin Yang\textsuperscript{\dag}}
\ead{yyb19@mails.tsinghua.edu}

\author[d]{Mingfeng Fan}
\ead{ming.fan@nus.edu.sg}

\author[e]{Haoran Wang}
\ead{wang_haoran@tongji.edu.cn}

\author[e]{Xiaocong Zhao\corref{cor1}}
\ead{zhaoxc@tongji.edu.cn}

\author[c]{Jianqiang Wang}
\ead{wjqlws@tsinghua.edu}

\address[a]{Singapore-MIT Alliance for Research and Technology (SMART), Singapore}
\address[b]{Department of Urban Studies and Planning, Massachusetts Institute of Technology, USA}
\address[c]{School of Vehicle and Mobility, Tsinghua University, China}
\address[d]{Department of Mechanical Engineering, National University of Singapore, Singapore}
\address[e]{Key Laboratory of Road and Traffic Engineering, Ministry of Education, Tongji University, China}


\cortext[cor1]{Corresponding author}
\fntext[fn2]{~\dag~ Heye Huang and Yibin Yang contributed equally to this work.}

%

\begin{abstract}
Safe autonomous driving in mixed traffic requires a unified understanding of multimodal interactions and dynamic planning under uncertainty. Existing learning-based methods often fail to capture rare but safety-critical behaviors, while rule-based systems lack adaptability in complex interactions. To address these limitations, we propose CogDrive, a cognition-driven multimodal prediction–planning fusion framework that integrates explicit modal reasoning with safety-aware decision optimization. The prediction module introduces cognitive representations of interaction modes based on topological motion semantics and nearest-neighbor relational encoding. By incorporating a differentiable modal loss and multimodal Gaussian decoding, CogDrive effectively learns sparse and unbalanced interaction behaviors, improving long-tail trajectory prediction accuracy. The planning module builds upon an emergency-response concept and develops a safety-stabilized trajectory tree optimization. Short-term consistent root trajectories ensure safety within replanning cycles, while long-term branches provide smooth and collision-free avoidance under low-probability or rapidly switching modes. Experiments on Argoverse2 and INTERACTION datasets show that CogDrive achieves state-of-the-art performance, reducing minADE and miss rate while maintaining smoothness. Closed-loop simulations further confirm stable and adaptive behavior across strong-interaction scenarios such as merging and intersections. By coupling cognitive multimodal prediction with safety-oriented planning, CogDrive establishes an interpretable and reliable paradigm for safe autonomy in complex traffic.

\end{abstract}


\end{frontmatter}


\section{Introduction}

In mixed traffic and real-world road environments, autonomous vehicles must inevitably interact with uncontrollable human-driven agents, giving rise to the challenge of \emph{interactive trajectory planning}. In such scenarios, the ego vehicle must generate dynamically feasible and safe trajectories in real time to pass through potential conflict regions~\citep{yang2024csdo, huang2024general}. Yet, this problem remains fundamentally paradoxical: overly conservative behavior leads to the ``freezing robot'' phenomenon, where vehicles stall and fail to progress, while overly aggressive strategies risk collisions and unsafe maneuvers.
This dilemma arises from two intertwined challenges. First, strongly interactive driving scenarios are sparse and imbalanced, making it difficult for learning-based methods to capture low-probability but safety-critical behaviors, often resulting in mispredictions at decisive moments~\citep{liu2025risknet}. Second, upstream trajectory predictions are inherently uncertain, exhibiting multiple possible modes that can shift abruptly or conflict with one another~\citep{ngiam2021scene}. Consequently, conventional planning methods struggle to account for all predicted modes in real time, frequently producing unstable or discontinuous trajectories. Since surrounding agents are uncontrollable, the ego vehicle cannot rely on centralized joint planning and must instead infer and respond to multimodal interactions in a decentralized and adaptive manner. However, this imbalance of multimodal behaviors often leaves the ego vehicle vulnerable to either collision due to prediction errors or excessive conservatism that results in inefficiency. The central question, therefore, is how to traverse such conflict zones both safely and efficiently.

Recent studies on motion forecasting and planning can be broadly categorized into two paradigms. Learning-based approaches exploit large-scale datasets and deep neural networks for behavior prediction~\citep{zhou2022hivt, jiang2023motiondiffuser}, 
but their purely data-driven nature limits interpretability and safety in rare or unseen conditions. In contrast, rule-based planners ensure physical feasibility and transparency~\citep{pek2020failsafe, ding2021epsilon}, 
yet often lack adaptability under highly interactive and uncertain traffic. 
Bridging these two paradigms calls for a unified framework that combines learning adaptability with rule-based safety reliability.

To address these limitations, this paper presents CogDrive, a cognition-driven multimodal prediction–planning fusion framework for safe autonomous driving. CogDrive integrates cognitive reasoning into the core decision process, coupling multimodal prediction with safety-stabilized planning within a unified structure. The framework learns to infer interaction semantics through differentiable modal reasoning while ensuring trajectory feasibility and stability under multimodal uncertainty~\citep{huang2025react}. By embedding cognition-inspired mechanisms, CogDrive transforms prediction and planning from two sequential stages into a coherent reasoning process that jointly perceives, anticipates, and acts. The main contributions of this paper are as follows:

\begin{itemize}
    \item We propose CogDrive, a cognition-driven multimodal prediction–planning fusion framework that unifies learning-based adaptability and rule-based safety stability within a unified cognitive decision process.    
    \item We design an interaction-aware multimodal prediction module, which encodes inter-agent semantics via modal classification and differentiable modal loss, improving the learning of sparse yet safety-critical behaviors.    
    \item We evaluate CogDrive on the Argoverse2 and INTERACTION datasets, achieving state-of-the-art performance in prediction accuracy, planning stability, and overall driving safety.
\end{itemize}

The remainder of this paper is organized as follows. Section 2 reviews related works on multimodal trajectory prediction and interactive planning. Sections 3,4 and 5 present the proposed CogDrive methodology, including cognitive prediction modeling and safety-stabilized planning. Section 6 reports experimental evaluations and comparisons with baselines. Section 6 concludes the paper and discusses future research directions.

%
\section{Related Works}
\label{rw}
%

Interactive single-vehicle trajectory planning methods fall into two broad categories: end-to-end and non-end-to-end ~\citep{cheng2024rethinking,chib2023recent}. End-to-end methods take raw sensor inputs and output trajectories or control commands via neural networks~\citep{prakash2021multi}. Recent advances in attention mechanisms, large-scale natural driving datasets, and improved architectures have substantially boosted performance~\citep{hwang2024emma}. However, these methods typically require high-fidelity sensors and realistic driver models in simulation, which limits their ability to capture multimodal uncertainty and creates a sim-to-real gap. Non-end-to-end methods address this by using detection and tracking to obtain states of other agents, combined with high-definition maps, to generate feasible trajectories for the ego vehicle. Non-end-to-end methods can also be divide into learning-based and rule-based planning.

\subsection{Learning-based Methods}

Learning-based planning methods are generally divided into two categories: imitation learning and reinforcement learning. Reinforcement learning explores policies through reward signals, but interaction strategies learned in simulation are often difficult to transfer to real-world driving due to the gap in fidelity and coverage. Imitation learning has also become a widely adopted and effective approach. Leveraging supervised learning on large-scale naturalistic driving datasets, it borrows techniques from trajectory and motion prediction to forecast ego trajectories, followed by post-processing to generate feasible and safe planning results. These methods can be further grouped into three categories discussed below.

\textbf{Embedding and Feature Representation.}  
Embedding methods map heterogeneous inputs, such as agent states and road centerlines, into latent representations that are crucial for interaction modeling and trajectory planning~\citep{jiang2025multi, shi2022motion}. Two main approaches are rasterized and vectorized embeddings. Rasterized inputs, typically derived from bird’s-eye-view images or LiDAR grids, enable fusion of diverse scene information but are computationally expensive and less effective for capturing long-range interactions. Their reliance on handcrafted feature extraction and limited receptive fields has further constrained their use in highly interactive scenarios.  
Vectorized embeddings have become the dominant paradigm by directly encoding agent trajectories and map elements as polylines. VectorNet pioneered this approach, leveraging graph neural networks to capture pairwise and group-level interactions among agents and road structures~\citep{gao2020vectornet}. Building on this, the MTR approach~\citep{sun2024controlmtr} applies PointNet~\citep{qi2017pointnet} to directly encode polylines with multilayer perceptrons (MLPs), followed by max-pooling to aggregate features as embedding representations. Such vectorized embedding methods have been widely adopted in vehicle trajectory prediction and have become a dominant and highly effective paradigm.

\textbf{Coordinate Systems and Normalization.}  
Coordinate system design is closely related to normalization, which stabilizes training by reducing gradient explosion or vanishing and improving optimization efficiency~\citep{ouyang2024deyo, carion2020end}. In single-agent prediction, ego-centric coordinates, where the ego vehicle’s position and heading define the frame, naturally provide translation and rotation invariance. This inductive bias allows the model to learn more efficiently and generally improves accuracy. Scene-centric coordinates, in contrast, align inputs to a global reference point, ensuring consistency but sacrificing flexibility for local interactions.  
For multi-agent prediction, unifying different agent frames is difficult. To address this, most recent work adopts an instance-centric design, where each agent or map element defines its own local frame. For example, road centerlines use midpoints and tangents, while vehicle trajectories use the current pose as the origin. This approach enables both normalization and symmetry across agents, while requiring mechanisms to handle transformations between frames~\citep{zhang2024simpl}. Representative methods include MTR++, which employs sinusoidal relative position encodings in attention layers~\citep{shi2022motion}, HPTR, which encodes relative coordinates with sine–cosine functions~\citep{zhang2023real}, and HiVT~\citep{zhou2022hierarchical} and QCNet~\citep{zhou2023query}, which construct local frames with rotation matrices to model inter-agent interactions. Instance-centric coordinates have thus become a robust compromise for interaction-aware trajectory prediction and planning.

\textbf{Encoder–Decoder Architectures and Multimodal Generation.}  
To address the multimodality inherent in interactive scenarios, a variety of encoder–decoder architectures have been proposed. Probabilistic generative methods, such as GANs~\citep{gupta2018social,eirale2025learning}, VAEs~\citep{salzmann2020trajectron++, cai2025mlig}, and diffusion models~\citep{shaoul2024multi,jiang2023motiondiffuser}, generate multiple candidate trajectories through sampling, but often suffer from mode collapse, limited interpretability, and high computational costs. Anchor-based approaches, exemplified by the TNT family~\citep{huang2020probabilistic}, predefine endpoints or high-level behaviors as anchors to produce trajectories with explicit modal distinctions, thereby mitigating mode compression. However, their performance is typically constrained by the trade-off between accuracy and the number of anchors.  
More recently, Transformer-based encoder–decoder frameworks with learnable query embeddings have emerged as a promising alternative for multimodal trajectory prediction~\citep{zhou2022hierarchical,zhou2023query}. These DETR-style architectures leverage attention mechanisms to jointly model spatial relations and behavioral diversity, offering improved scalability and interpretability compared to sampling- or anchor-based approaches. Each learnable query represents a potential behavioral mode, allowing the network to capture both common and rare interactions within a unified attention space. By replacing stochastic sampling with deterministic relational reasoning, these query-driven architectures achieve higher stability, controllable diversity, and better alignment between predicted modes and real driving behaviors, making them one of the most advanced paradigms for learning-based multimodal prediction.

\subsection{Rule-based Methods}
In contrast, rule-based methods emphasize interpretability and safety, typically generating trajectories through predefined models or rules. The core challenge lies in making safe decisions under uncertainty.
\textbf{(a) Maximum Likelihood Planning.}  
These methods assume surrounding agents will follow their most probable behaviors and plan ego trajectories accordingly. Methods include Monte Carlo tree search~\citep{lenz2016tactical}, finite-state machines~\citep{meng2021fsm}, raster or optimizations~\citep{huang2025lead}. They are computationally efficient but often fail to handle rare yet dangerous behaviors, which can result in discontinuous or unsafe decisions under unexpected interactions.
\textbf{(b) Partially Observable Markov Decision Processes.}  
POMDP frameworks explicitly model uncertainty and have been applied in interactive planning systems such as EPSILON, which integrates behavior planning with optimization-based motion planning~\citep{ding2021epsilon,sheng2024safe}. EPSILON employs guided branching in action–observation spaces and a spatio-temporal semantic corridor to generate safe, smooth trajectories. These methods provide clear interpretability under uncertainty but remain computationally demanding in large-scale dynamic traffic.
\textbf{(c) Defensive and Contingency Planning.}  
Defensive planning approaches generate trajectory trees with shared initial segments and branches to hedge against different predicted futures~\citep{huang2024general}. Fail-safe motion planning or contingency MPC generate such trees to cover multiple modalities ~\citep{pek2020failsafe}. They ensure short-term safety but scale poorly: trajectory tree size grows exponentially with prediction horizon and number of agents, limiting real-time use. They provide strong safety guarantees and interpretability, but often lack adaptivity and flexibility when confronted with highly interactive, multimodal uncertainties.


\section{Framework Overview}
\label{sec:framework}

CogDrive follows the principle of cognition-driven autonomy, where cognitive mechanisms bridge perception, reasoning, and control. It enables accurate modeling and digital representation of the human–vehicle–road system, capturing the intrinsic properties, interactions, and governing dynamics of each element. By inheriting the interpretability of rule-based mechanisms and the adaptability of data-driven learning, CogDrive empowers autonomous systems with generalization, evolution, and reliable decision-making capabilities. Building on this foundation, CogDrive formulates interaction-aware motion generation in mixed traffic as a unified process that couples multi-agent trajectory prediction with ego-vehicle planning. This dual-stage structure allows the system to anticipate multimodal interaction outcomes while generating dynamically feasible and safety-consistent plans.

\textbf{Multi-Agent Joint Trajectory Prediction.} In complex traffic, each agent’s future motion depends on both its own history and the behaviors of surrounding participants. Given the observed trajectories $S_A \in \mathbb{R}^{M \times T_h \times C_a}$ and HD map data $S_R \in \mathbb{R}^{N_r \times N_p \times C_r}$, the prediction module estimates the joint distribution of future trajectories:
\begin{equation}
P(Y|S_A,S_R),
\end{equation}
where $M$ is the number of dynamic agents and $(N_r, N_p, C_r)$ describe lane features. Each agent’s motion is modeled under multiple behavioral modes using a Gaussian Mixture Model (GMM):
\begin{equation}
P(Y^k_i|S_A,S_R)=\mathcal{N}(\mu^k_{i,t},\Sigma^k_{i,t}),
\end{equation}
and the overall predictive distribution is
\begin{equation}
P(Y_{i,t}|S_A,S_R)=\sum_{k=1}^{K} p_k P(Y^k_{i,t}|S_A,S_R),
\end{equation}
where $\sum_{k=1}^K p_k=1$. This probabilistic formulation captures multimodal uncertainty and enables reasoning over rare but safety-critical behaviors.

\textbf{Ego-Vehicle Safety-Stabilized Trajectory Planning.} Based on the multimodal prediction $P(Y|S_A,S_R)$, the planner generates a feasible and safe ego trajectory $\mathcal{T}$. Unlike conventional one-shot planners, CogDrive adopts a cognition-inspired emergency-aware approach. It first constructs a root trajectory $\mathcal{T}^{root}_{0:T_b}$ ensuring short-term safety, then extends branched trajectories $\mathcal{T}^{k}_{T_b+1:T}$ for different modes. The branching time $T_b$ exceeds the replanning cycle, ensuring collision-free, dynamically feasible execution.
This hierarchical design couples prediction and planning bidirectionally: multimodal predictions provide interpretable intent cues, while the planner ensures safe realization of each mode through adaptive optimization. Together, they form the cognitive backbone of CogDrive, enabling safe and explainable autonomy in complex mixed traffic.

\section{Multimodal Joint Trajectory Prediction}
\label{methodology1}
%

\begin{figure}[!t]
\centering
\includegraphics[width=\linewidth]{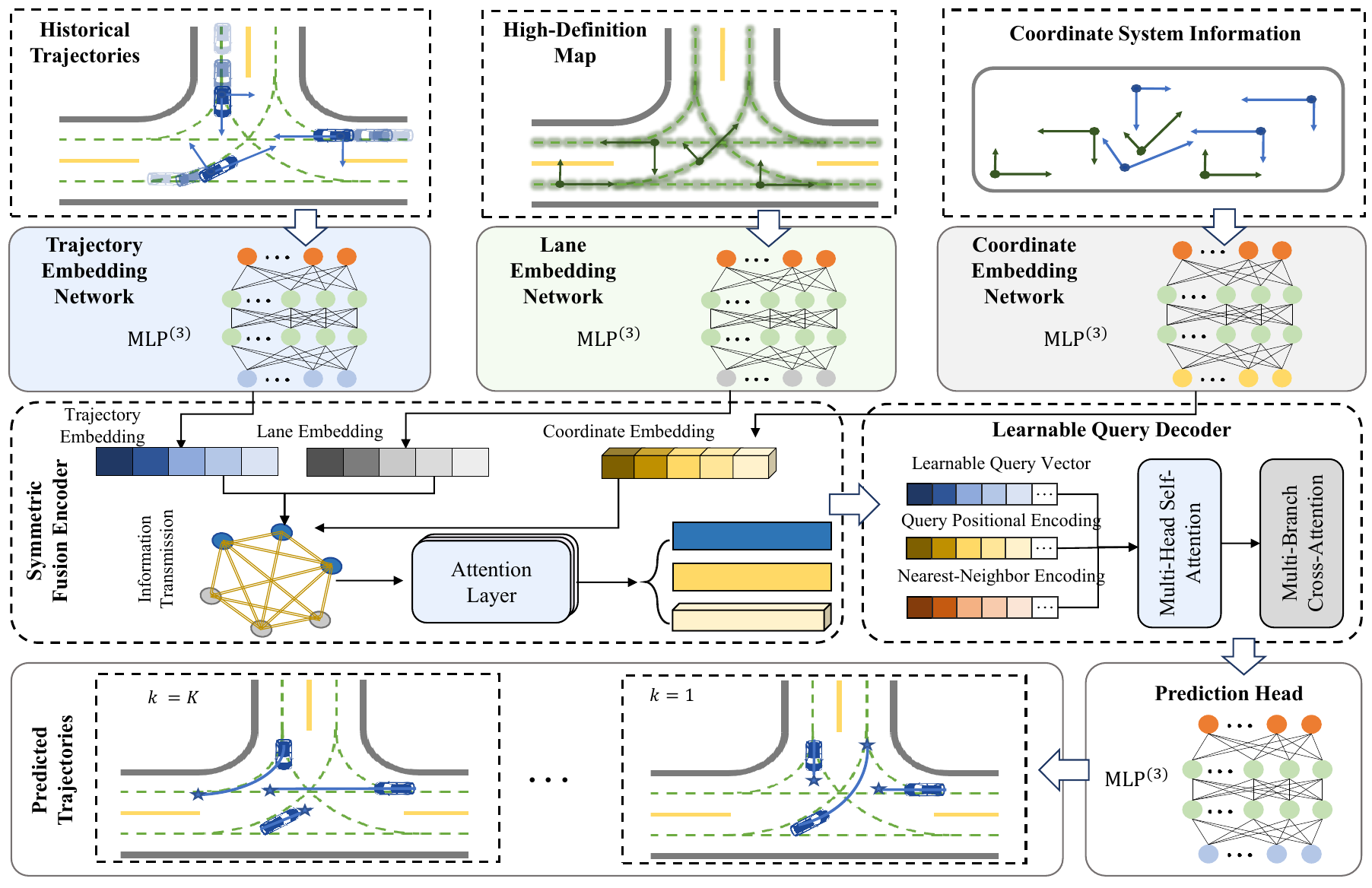}
\caption{Overview of the cognition-driven multimodal prediction network in \textsc{CogDrive}. Historical trajectories, high-definition maps, and local coordinate information are encoded through three MLP-based embedding networks. Their outputs are fused by a symmetric fusion encoder that models pairwise spatial and behavioral relations via relative positional and nearest-neighbor encoding. Learnable query decoding with multi-branch cross-attention generates multimodal joint trajectories, each representing a distinct interaction mode between the ego and surrounding agents.}
\label{fig:framework_prediction}
\end{figure}

The proposed CogDrive framework formulates multimodal joint trajectory prediction as a cognition-driven learning process, enabling interpretable and accurate reasoning of motion intentions among interacting agents. As illustrated in Fig.~\ref{fig:framework_prediction}, the network integrates scene geometry, agent dynamics, and coordinate information within a unified vectorized representation to model diverse behavioral modalities.
Each observed trajectory and high-definition (HD) map segment is first transformed into a local instance-centric coordinate system to preserve geometric consistency across heterogeneous agents. For dynamic agents, the current state defines the origin and heading of its local frame, while for static map elements, the centroid and lane orientation define the axes. This instance-level normalization ensures translation–rotation invariance and facilitates consistent spatial reasoning across different traffic configurations.
Three embedding networks are implemented to encode trajectory, lane, and coordinate features using multilayer perceptrons (MLPs). These embeddings are then fused through a \emph{symmetric fusion encoder} that captures pairwise interactions between agents and map elements via relative positional encoding and nearest-neighbor attention. This design supports bidirectional information exchange, allowing each entity to reason jointly about spatial and behavioral relations in the traffic scene.

In the decoding stage, a set of learnable query vectors interacts with the encoded context through multi-head attention and cross-attention layers. Each query corresponds to a distinct behavioral mode, such as yielding, merging, or accelerating, and evolves iteratively to generate a trajectory hypothesis with an associated probability. The decoder’s multi-branch structure enables controllable multimodal output, while query positional and modality encodings ensure interpretability of each interaction mode. 
The resulting prediction head produces a probabilistic distribution of multimodal joint trajectories, where each branch captures a physically plausible motion pattern between the ego and surrounding agents. This cognition-driven design allows \textsc{CogDrive} to represent both common and rare interaction behaviors, bridging perception and decision modules through interpretable multimodal reasoning.


\subsection{Behavioral Modality Modeling}
\label{sec:modality}

To represent diverse and controllable interaction patterns among agents, CogDrive introduces a cognitive behavioral modality representation inspired by topological motion equivalence. Traditional single-vehicle planning assumes that trajectories with identical start and end points can be smoothly deformed without crossing obstacles, forming a topological homotopy class. Extending this idea to interactive driving, CogDrive encodes agent-to-agent behaviors using continuous deformation relationships that distinguish different motion intentions, such as yielding, merging, or overtaking. 

Given two agents $i$ and $j$ with trajectories $\mathcal{T}_i$ and $\mathcal{T}_j$, the relative angular displacement between them is quantified by the cumulative change in their relative bearing:
\begin{equation}
\Delta \theta_m(\mathcal{T}_i,\mathcal{T}_j)=\sum_{t=0}^{T-1} f_{norm}\left( \arctan\frac{y_{i}^{t+1}-y_{j}^{t+1}}{x_{i}^{t+1}-x_{j}^{t+1}} - \arctan\frac{y_{i}^{t}-y_{j}^{t}}{x_{i}^{t}-x_{j}^{t}} \right),
\end{equation}
where $f_{norm}(\cdot)$ normalizes the angular difference to the interval $[-\pi,\pi]$. By applying a threshold $\hat{\theta}$, interaction modes are categorized into three discrete types:
\begin{equation}
m(\mathcal{T}_i,\mathcal{T}_j)=
\begin{cases}
-1, & \Delta \theta_m < -\hat{\theta},\\
0, & -\hat{\theta} \leq \Delta \theta_m \leq \hat{\theta},\\
1, & \Delta \theta_m > \hat{\theta}.
\end{cases}
\end{equation}
Here, $m=-1$, $0$, and $1$ respectively represent yielding, neutral, and aggressive behaviors between two vehicles. This compact encoding allows the network to classify and predict interaction modes directly from spatial relationships.

In multi-agent settings, the ego vehicle’s modality vector is constructed as
\begin{equation}
\mathbf{m}(\mathcal{T}_{ego},\mathcal{T}_1,\dots,\mathcal{T}_{M-1}) = [m(\mathcal{T}_{ego},\mathcal{T}_1),\dots,m(\mathcal{T}_{ego},\mathcal{T}_{M-1})],
\end{equation}
where $M$ denotes the number of surrounding agents. The modality vector compactly represents the ego’s behavioral relationship with multiple neighbors, effectively capturing the combinatorial nature of interactive driving. To avoid combinatorial explosion, CogDrive considers only the nearest neighbors in interaction space, enabling tractable learning and real-time inference. During training, a differentiable approximation of the modality function is used to enable gradient-based optimization, allowing the network to learn smooth and interpretable transitions between interaction behaviors. This formulation provides the basis for controllable and cognitively grounded multimodal trajectory prediction in strongly interactive environments.


\subsection{Scene Representation and Relative Feature Encoding}
\label{sec:feature}

To ensure geometric consistency and spatial invariance in multimodal trajectory prediction, CogDrive employs an instance-centric coordinate representation. For each predicted agent, a local coordinate frame is established with its current position as the origin and its heading direction aligned with the $x$-axis. Other scene elements, including surrounding agents and lane polylines, are transformed from the global map frame into this local coordinate system. This normalization not only preserves the spatial structure of the environment but also significantly improves the efficiency and generalization of the learning process.

However, in multi-agent prediction tasks, each agent resides in its own coordinate space, leading to a loss of mutual geometric reference. To recover this information, CogDrive introduces relative positional encoding that models pairwise spatial relationships among all instances. Let $\mathbf{z}_i=(x_i,y_i)$ denote the origin of the local coordinate system for instance $i$, and $\theta_i$ its heading angle. For any pair of instances $(i,j)$, their relative orientation and distance are computed as
\begin{equation}
\theta_{i,j} = \theta_j - \theta_i, \quad
\beta_{i,j} = \arctan\frac{y_j - y_i}{x_j - x_i} - \theta_i, \quad
d_{i,j} = \|\mathbf{z}_i - \mathbf{z}_j\|.
\end{equation}

These parameters represent the heading difference, bearing angle, and distance between two instances. The relative positional feature $\mathbf{r}_{p,i,j}$ is defined as a five-dimensional vector:
\begin{equation}
\mathbf{r}_{p,i,j} = [\sin \theta_{i,j}, \cos \theta_{i,j}, \sin \beta_{i,j}, \cos \beta_{i,j}, d_{i,j}].
\end{equation}

This representation ensures rotational invariance and smooth continuity, enabling the model to learn interaction patterns independent of absolute positions. 
All pairwise features are concatenated into a tensor $\mathbf{R}_p \in \mathbb{R}^{N\times N\times5}$ and fed into the attention-based fusion encoder. 
By embedding scene geometry in this relational form, CogDrive achieves compact and interpretable spatial reasoning across agents and road elements.

\begin{figure}[!t]
\centering
\includegraphics[width=\linewidth]{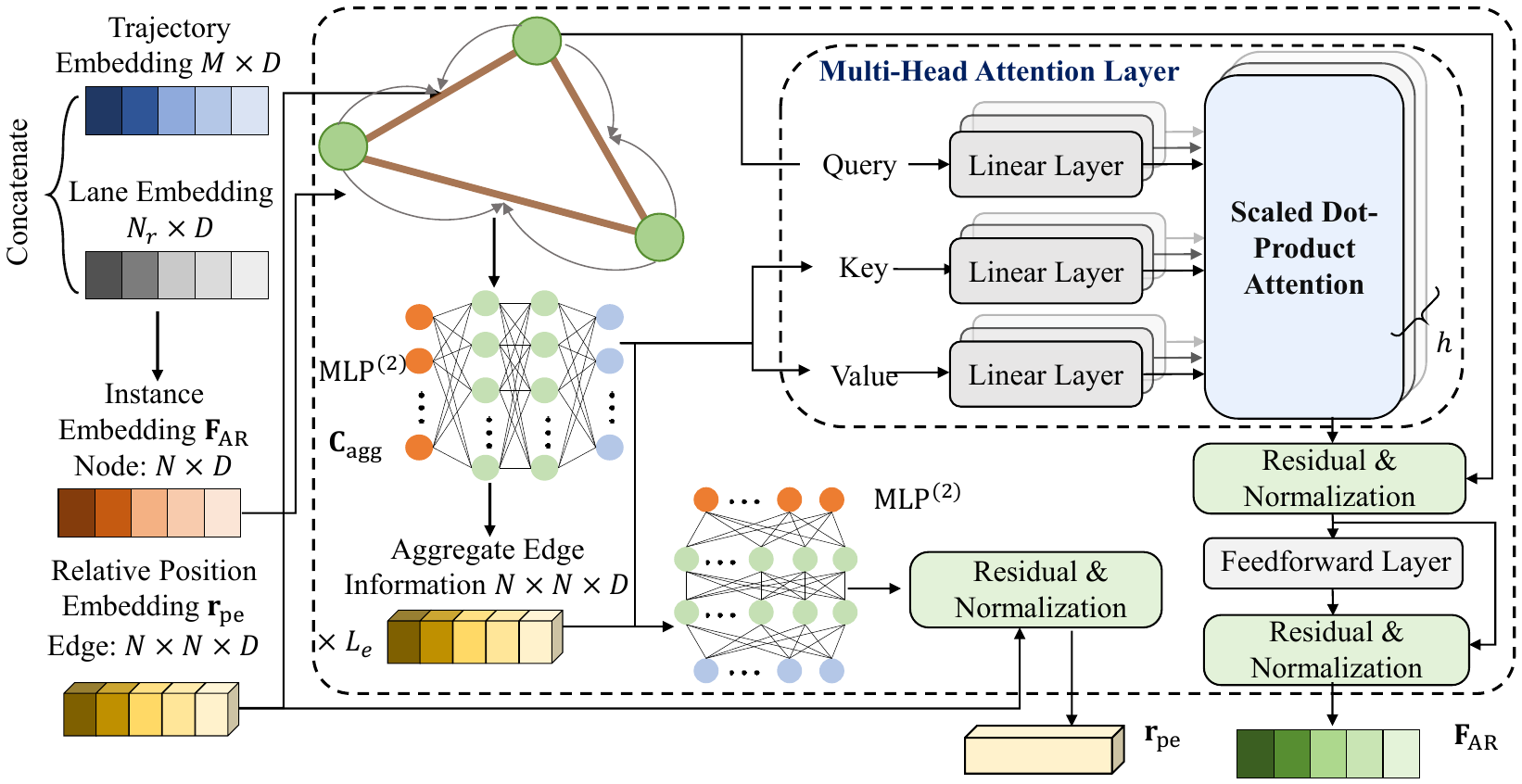}
\caption{Architecture of the symmetric fusion encoder. Its structure resembles a self-attention model but explicitly introduces relational encoding between different instance-centric coordinate systems. Through symmetric feature updates and relative positional embeddings, the encoder preserves viewpoint and ordering invariance across instances. Each instance is represented as a node, and their pairwise coordinate transformations define directed edges, forming a fully connected self-looped graph that ensures consistent bidirectional fusion of multimodal features.}
\label{fig:encoder}
\end{figure}

%
\subsection{Vectorized Embedding and Symmetric Context Encoding}
\label{sec:embedding}

At the input stage, CogDrive transforms heterogeneous scene elements, including historical trajectories $S_A$, lane segments $S_R$, and relative positional features $\mathbf{r}_p$, into a unified latent space through vectorized embedding. Inspired by VectorNet and PointNet architectures, each trajectory polyline or map segment is first transformed to its corresponding instance-centric coordinate system, followed by vector-wise feature extraction using multilayer perceptrons (MLPs). The embedding process can be written as
\begin{equation}
\mathbf{F}_A = \rho(\text{MLP}(\Phi(S_A))), \qquad
\mathbf{F}_R = \rho(\text{MLP}(\Phi(S_R))), \qquad
\mathbf{r}_{pe} = \text{MLP}(\mathbf{r}_p),
\end{equation}
where $\Phi$ denotes coordinate transformation and $\rho(\cdot)$ represents max-pooling along the temporal dimension. This design yields compact trajectory features $\mathbf{F}_A \in \mathbb{R}^{M \times D}$, lane features $\mathbf{F}_R \in \mathbb{R}^{N \times D}$, and relative positional embeddings $\mathbf{r}_{pe} \in \mathbb{R}^{N \times N \times D}$ that preserve both spatial topology and motion continuity.

The encoded features are concatenated as $\mathbf{F}_{AR}=[\mathbf{F}_A,\mathbf{F}_R]$ and passed to a symmetric fusion encoder based on a Transformer architecture, as illustrated in Fig.~\ref{fig:encoder}. Unlike conventional attention mechanisms, the symmetric fusion encoder explicitly maintains reciprocal relationships among different instance-centric coordinate systems. For each pair of instances $(i,j)$, the model aggregates directional and relational features using MLPs:
\begin{equation}
\mathbf{C}'_{agg,i,j} = \text{MLP}(\text{Concat}(\mathbf{F}'_{AR,i},\mathbf{F}'_{AR,j},\mathbf{r}'_{pe,i,j})),
\end{equation}
where $\mathbf{C}'_{agg}$ encodes bidirectional contextual dependencies. A multi-head attention (MHA) module then updates the feature representation:
\begin{equation}
\mathbf{F}'^{\,l+1}_{AR} = \text{MHA}(\mathbf{F}'^{\,l}_{AR}, \mathbf{C}'_{agg}, \mathbf{C}'_{agg}),
\end{equation}
allowing information to propagate symmetrically across agents and map segments. Residual connections and layer normalization preserve stability and gradient flow, while additional MLP layers refine relative embeddings:
\begin{equation}
\mathbf{r}'^{\,l+1}_{pe} = \text{MLP}(\mathbf{C}'_{agg}) + \mathbf{r}'^{\,l}_{pe}.
\end{equation}

After $L_e$ layers of iterative encoding, the network outputs updated relational features $\mathbf{F}'_{AR}$ and $\mathbf{r}'_{pe}$ that capture bidirectional spatial dependencies and cross-coordinate consistency. This symmetric relational design ensures that all pairwise interactions are represented in an order-invariant and geometrically consistent manner, enabling CogDrive to reason over complex inter-agent dependencies. The fused relational features $\mathbf{F}'_{AR}$ are subsequently used in the decoding stage for multimodal trajectory generation and interaction-aware risk prediction.

\begin{figure}[!t]
\centering
\includegraphics[width=\linewidth]{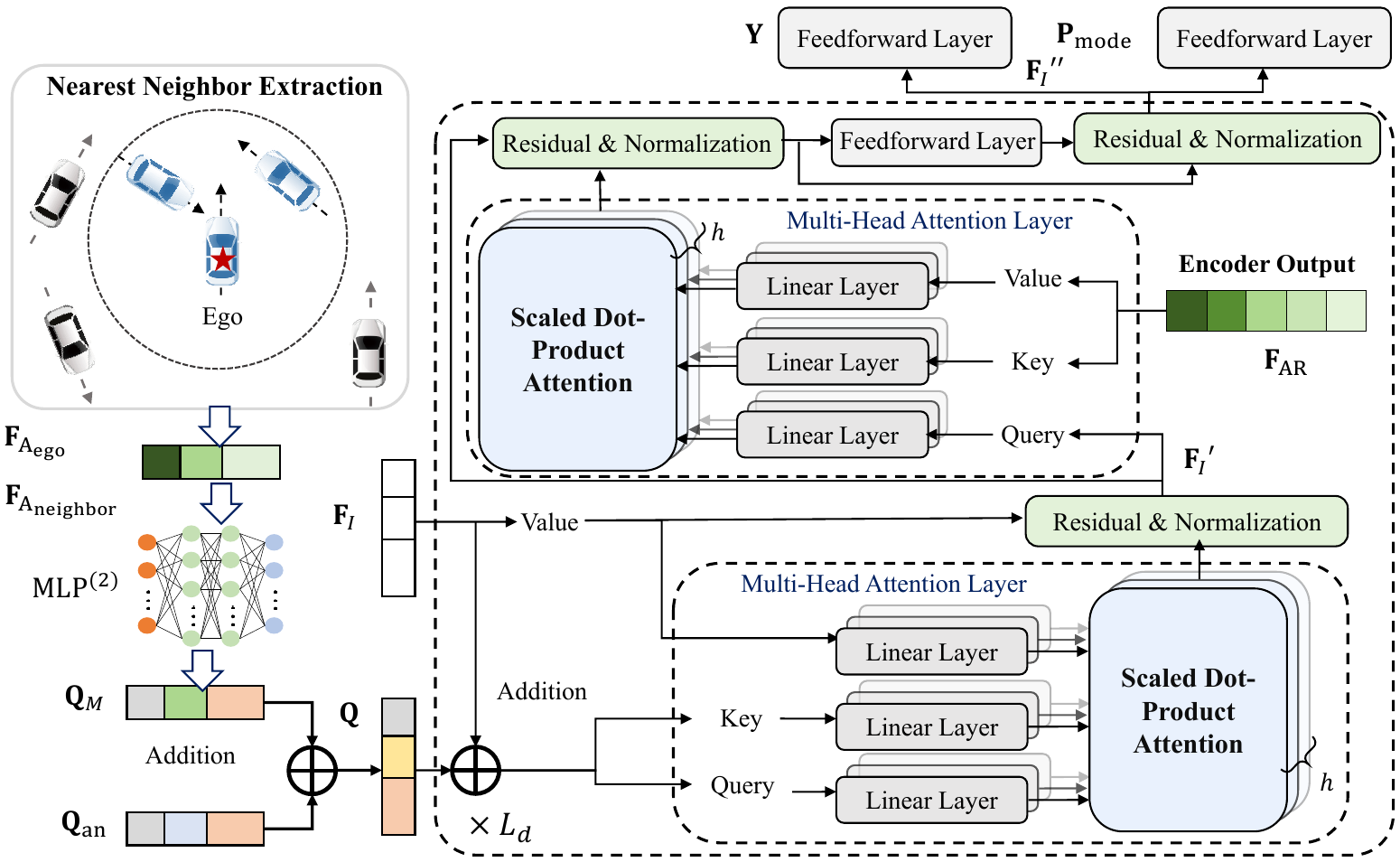}
\caption{Decoder architecture for interaction-aware representation learning. The decoder receives relational features from the symmetric fusion encoder and generates multimodal hypotheses through iterative self- and cross-attention. Each learnable query combines an anchor component and a modality-guided component derived from the ego and its nearest neighbors, enabling interpretable reasoning over distinct behavioral modes. The decoder progressively refines these queries into trajectory hypotheses with associated probabilities, forming a complete mapping from interaction context to multimodal motion prediction.}
\label{fig:decoder}
\end{figure}


\subsection{Learnable Decoding}
\label{sec:decoding}

To achieve multimodal and interpretable trajectory prediction, CogDrive adopts a learnable query-based decoding mechanism inspired by the DETR architecture. As illustrated in Fig.~\ref{fig:decoder}, the decoder generates a set of learnable queries in latent space and associates them with behavioral mode embeddings that represent distinct interaction intentions, such as yielding, merging, or overtaking. Each query interacts with encoded contextual features through multi-head and cross-attention layers, gradually refining its spatial hypothesis into a trajectory mode with corresponding probability.

The query embedding $\mathbf{Q}$ is composed of two components: a learnable anchor query $\mathbf{Q}_{an}$ and a modality-guided query $\mathbf{Q}_M$. The modality-guided component is generated from the encoded features of the ego vehicle and its selected neighboring agents:
\begin{equation}
\mathbf{Q}_M = \text{MLP}\left([\mathbf{F}'_{ego},\mathbf{F}'_{neighbor}]\right),
\end{equation}
where $\mathbf{F}'_{ego}$ and $\mathbf{F}'_{neighbor}$ denote the updated latent representations from the symmetric fusion encoder. To limit computational complexity, only the $n_{neighbor}$ most relevant agents are considered based on proximity or potential collision risk. The total query embedding is defined as
\begin{equation}
\mathbf{Q} = \mathbf{Q}_{an} + \mathbf{Q}_M,
\end{equation}
where $\mathbf{Q}\in\mathbb{R}^{k_M\times D}$ contains $k_M$ learnable queries, each representing a potential interaction mode.

Within each decoding layer, self-attention ensures the diversity of query embeddings, preventing mode collapse, while cross-attention enables each query to interact with the encoded feature map $\mathbf{F}_I$. The computation process follows
\begin{equation}
\mathbf{F}'_I = \text{MHA}(\mathbf{F}_I + \mathbf{Q},\, \mathbf{F}_I + \mathbf{Q},\, \mathbf{F}_I),
\end{equation}
where MHA denotes the standard multi-head attention operation. Subsequently, cross-attention aggregates contextual information between the query features and the encoder output $\mathbf{F}'_{AR}$:
\begin{equation}
\mathbf{F}''_I = \text{MHA}(\mathbf{F}'_I + \mathbf{Q},\, \mathbf{F}'_{AR},\, \mathbf{F}'_{AR}).
\end{equation}

After $L_d$ decoding layers, the final output $\mathbf{F}''_I$ represents multimodal latent trajectories, which are projected through two independent fully connected layers to obtain the predicted trajectory coordinates $X$ and the mode probability vector
\begin{equation}
\mathbf{P}_{mode} = (p_1, p_2, \dots, p_K).
\end{equation}

This learnable decoding mechanism allows CogDrive to autonomously discover and refine distinct interaction patterns through iterative attention updates, bridging the gap between cognition-driven behavior understanding and accurate multimodal trajectory forecasting. It further ensures that rare yet safety-critical interaction modes remain represented, supporting robust downstream planning and decision-making.


\subsection{Network Training}
\label{sec:training}

CogDrive is trained under a Winner-Takes-All (WTA) strategy that jointly optimizes trajectory regression, classification, and modality consistency. The overall training objective combines three complementary losses:
\begin{equation}
\mathcal{L} = \mathcal{L}_{reg} + \alpha_1 \mathcal{L}_{cls} + \alpha_2 \mathcal{L}_{mode},
\end{equation}
where $\alpha_1$ and $\alpha_2$ are balancing coefficients set to 0.2 and 0.01, respectively. The WTA scheme selects the predicted mode $k^*$ with the minimum final displacement error and updates the corresponding mode probability and trajectory through backpropagation.

For multimodal classification, CogDrive employs a max-margin loss to ensure the separability of different modes:
\begin{equation}
\mathcal{L}_{cls} = \sum_{k=1}^{K} \max(0, \epsilon_{margin} + p_k - p_{k^*}),
\end{equation}
where $p_k$ is the probability of mode $k$, $p_{k^*}$ corresponds to the optimal mode, and $\epsilon_{margin}$ is a predefined margin (set to 0.2 in experiments). This margin-based constraint prevents mode collapse by maintaining sufficient separation between mode probabilities, encouraging the network to learn diverse interaction hypotheses.

The regression loss refines trajectory accuracy and motion smoothness through position and yaw-angle supervision:
\begin{equation}
\mathcal{L}_{reg} = \mathcal{L}_{pos}(\bar{Y}_{pos}, Y^*_{pos}) + \mathcal{L}_{yaw}(\bar{Y}_{yaw}, Y^*_{yaw}),
\end{equation}
where $\bar{Y}$ denotes predicted trajectories of the best mode $k^*$, and $Y^*$ represents ground-truth references. The positional loss $\mathcal{L}_{pos}$ is computed as the mean squared error of predicted coordinates, while the yaw loss captures angular consistency between predicted and true heading directions:
\begin{equation}
\mathcal{L}_{yaw}(\bar{Y}_{yaw}, Y^*_{yaw}) = \frac{1 - \phi_{CosSim}(\bar{Y}_{yaw}, Y^*_{yaw})}{2},
\end{equation}
where $\phi_{CosSim}$ measures cosine similarity between yaw vectors. This design penalizes orientation discontinuities and promotes physically plausible trajectories, especially under low-speed or near-collision scenarios where continuous steering control is critical.

For multimodal regularization, CogDrive introduces a differentiable surrogate to approximate the discrete mode-matching function, enabling end-to-end optimization of mode consistency. The resulting $\mathcal{L}_{mode}$ term measures pairwise alignment between predicted and reference modes while maintaining smooth gradient propagation. This improves both multimodal coverage and inter-mode calibration, ensuring that the learned modes correspond to physically interpretable driving behaviors.
By combining position, yaw, and mode-aware objectives, CogDrive achieves a balanced optimization between accuracy and diversity. The model learns to precisely predict future trajectories while maintaining multimodal separability, producing realistic, physically consistent, and interaction-aware motion forecasts across diverse and complex traffic scenarios.

\section{Multimodal Safety-Aware Trajectory Planning}
\label{sec:planning}
%
\begin{figure}[!t]
\centering
\includegraphics[width=1\linewidth]{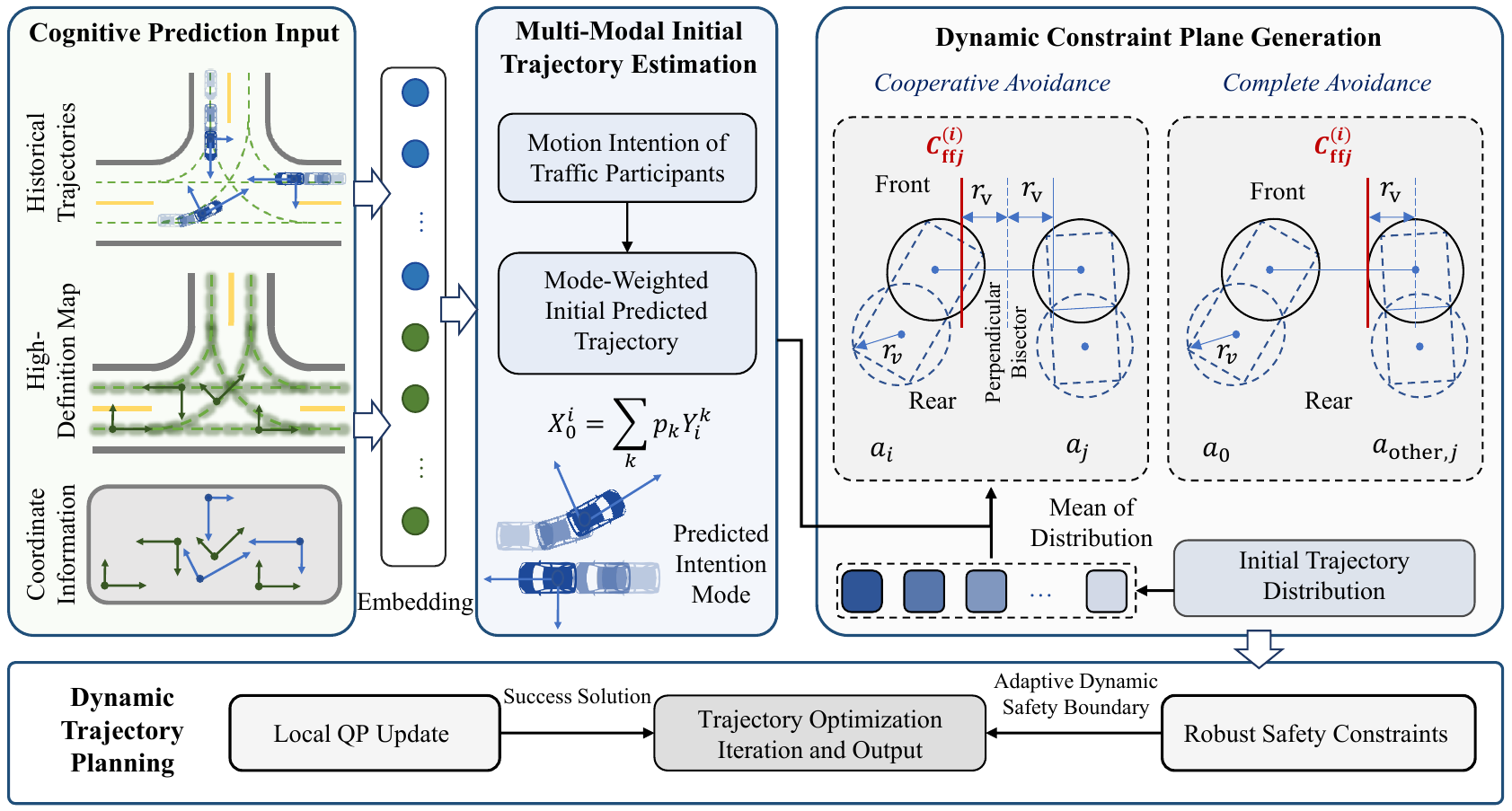}
\caption{Dynamic safety-aware trajectory planning in CogDrive. Cognitive prediction provides a mode-weighted nominal trajectory, from which dynamic constraint planes and an adaptive safety boundary are constructed. Through local QP updates, the planner yields collision-free solutions that realize either cooperative yielding or complete avoidance, aligning the ego motion with multimodal interaction intentions.}
\label{fig:planning}
\end{figure}

The multimodal trajectory prediction module generates $K$ probabilistic trajectory hypotheses $\mathbf{Y}$ representing distinct behavioral intentions of surrounding agents. Building upon these predictions, CogDrive performs safety-aware emergency trajectory tree planning to ensure short-term collision avoidance and long-term behavioral stability under multimodal uncertainty. The framework integrates multimodal preparedness planning with single-vehicle trajectory optimization into a unified hierarchical process, as illustrated in Fig.~\ref{fig:planning}. The planner (i) constructs a mode-weighted nominal trajectory as initialization, (ii) generates dynamic constraint planes between ego and neighbors, and (iii) maintains an adaptive safety boundary that is updated via local quadratic programming (QP) iterations to realize cooperative or complete avoidance consistent with multimodal intents.

\subsection{Multimodal Preparedness Planning}
\label{sec:preparedness}

When interacting with uncertain agents, the ego vehicle must maintain safety across all possible motion modes. To achieve this, CogDrive adopts a preparedness-based planning strategy that explicitly accounts for the distribution of predicted behaviors. Instead of following only the most probable mode, the planner generates a short-horizon emergency trajectory that guarantees safety across all modes, and a long-horizon trajectory that maintains continuity after mode resolution. This design mitigates overreactive behavior and ensures decision consistency during interaction. As indicated in Fig.~\ref{fig:planning}, the emergency branch covers the full set of predicted modes, while the nominal branch tracks the most consistent mode as the belief evolves.

The multimodal emergency planning problem is formulated as a constrained nonlinear optimization:
\begin{equation}
\min_{\mathbf{U}} 
\sum_{t=0}^{T_b} l_z(\mathbf{z}_t) + \sum_{t=1}^{T_b} l_u(\mathbf{u}_t)
+ \sum_{k \in K}\sum_{t=T_b}^{T} \big(l_z(\mathbf{z}^k_t) + l_u(\mathbf{u}^k_t)\big),
\end{equation}
subject to
\begin{align}
\mathbf{z}_t &= f(\mathbf{z}_{t-1}, \mathbf{u}_t), && \forall t \in [1, T_b], \\
\mathbf{z}^k_t &= f(\mathbf{z}^k_{t-1}, \mathbf{u}^k_t), && \forall t \in (T_b, T], \, \forall k \in [1, K], \\
h_i(\mathbf{z}_t) &\le 0, && \forall t \in [0, T_b], \\
h^k_i(\mathbf{z}^k_t) &\le 0, && \forall t \in (T_b, T], \, \forall k \in [1, K],
\end{align}
where $\mathbf{z}_t \in \mathbb{R}^{n_z}$ and $\mathbf{u}_t \in \mathbb{R}^{n_u}$ denote the ego state and control input at time step $t$, $f(\cdot)$ represents discrete-time vehicle dynamics, and $h_i(\cdot)$ defines the safety and comfort constraints. The cost functions $l_z(\cdot)$ and $l_u(\cdot)$ penalize deviations from desired states and excessive control efforts. The first phase $[0, T_b]$ ensures immediate safety across all possible interactions, while the second phase $(T_b, T]$ maintains long-term stability and smoothness.

\subsection{Single-Vehicle Trajectory Planning}
\label{sec:singleplanning}

Building on the multimodal preparedness strategy, CogDrive further refines the ego-level planning process by explicitly formulating a geometric and optimization-based single-vehicle trajectory planner. This planner bridges the cognition-driven predictions and low-level control through a unified optimization procedure that enforces dynamic feasibility, spatial safety, and temporal smoothness.

\textbf{Initialization from multimodal predictions.}
The planner receives $K$ multimodal trajectory hypotheses $\{Y^k_i\}_{k=1}^{K}$ from the prediction module, where each $Y^k_i = [y^k_{i,0}, \ldots, y^k_{i,T}]$ represents the predicted future motion of agent $i$ under interaction mode $k$, and $p_k$ denotes the probability of that mode. The ego vehicle constructs a weighted nominal trajectory
\begin{equation}
\bar{\mathbf{X}}_0 = \sum_{k=1}^{K} p_k\, Y^k_{\mathrm{ego}},
\end{equation}
and generates a set of candidate trajectories $\{\bar{\mathbf{X}}_0^m\}_{m=1}^{M}$ by applying small perturbations around $\bar{\mathbf{X}}_0$. These trajectories serve as adaptive initial guesses for subsequent optimization, ensuring consistency between prediction and planning. As shown in Fig.~\ref{fig:planning}, this step corresponds to the mode-weighted initial predicted trajectory.

\textbf{Dynamic geometric constraint generation.}
Since surrounding vehicles are uncontrollable, CogDrive models their interactions with the ego vehicle through dynamically updated geometric constraints. Each vehicle is approximated by front and rear circular envelopes with radii $r_F$ and $r_R$, respectively, defining its physical occupancy. For a neighboring vehicle $j$ relative to the ego vehicle $i$, a separating hyperplane is constructed using their center displacement vector $\mathbf{d}_{ij} = \mathbf{p}_j - \mathbf{p}_i$, where $\mathbf{p}_i = [x_i, y_i]^\top$ and $\mathbf{p}_j = [x_j, y_j]^\top$ are their planar positions. The linear constraint is expressed as
\begin{equation}
\mathbf{A}_{c,ij}\mathbf{Y}_{c,i} \le \mathbf{b}_{c,ij},
\quad 
\mathbf{A}_{c,ij} = \frac{\mathbf{d}_{ij}^\top}{\|\mathbf{d}_{ij}\|}, 
\quad
\mathbf{b}_{c,ij} = \|\mathbf{d}_{ij}\| - (r_F + r_R),
\end{equation}
where $\mathbf{Y}_{c,i} = [x_{F,i}, y_{F,i}, x_{R,i}, y_{R,i}]^\top$ denotes the concatenated coordinates of the front and rear wheel centers of the ego vehicle. This constraint enforces a minimum separation distance between the ego and neighboring vehicles, thus defining a time-varying safe region. In Fig.~\ref{fig:planning}, the perpendicular bisector visualization clarifies the construction of dynamic constraint planes, from which cooperative yielding or complete avoidance behaviors naturally emerge under different multimodal hypotheses.

\textbf{Static and robust corridor constraints.}
To account for environmental boundaries and uncertainty in multimodal predictions, CogDrive constructs a robust safety corridor around static obstacles and lane boundaries. Each corridor is represented as a convex polytope that bounds the ego trajectory within upper and lower limits:
\begin{equation}
\mathbf{Y}_{c,\min} \le \mathbf{Y}_{c,t+1} \le \mathbf{Y}_{c,\max}, 
\end{equation}
where $\mathbf{Y}_{c,\min}$ and $\mathbf{Y}_{c,\max}$ are adaptively expanded according to the predicted positional uncertainty $\Sigma^k_{\mathrm{ego}}$ from the multimodal prediction module. This constraint ensures that even under bounded perception or model errors, the resulting trajectory remains collision-free and dynamically feasible. The evolving adaptive dynamic safety boundary in Fig.~\ref{fig:planning} illustrates this mechanism.

\textbf{Quadratic programming formulation.}
Integrating the dynamic and static constraints, the ego trajectory optimization is formulated as a constrained quadratic program:
\begin{equation}
\min_{\mathbf{X},\mathbf{U}} 
\sum_{t=0}^{T} 
\|\mathbf{X}_t - \bar{\mathbf{X}}_t\|_{Q}^{2} + 
\|\mathbf{U}_t - \bar{\mathbf{U}}_t\|_{R}^{2},
\end{equation}
subject to
\begin{align}
\mathbf{X}_{t+1} &= f_d(\mathbf{X}_t, \mathbf{U}_t), && \forall t \in [0, T], \\
\mathbf{A}_{c,ij}\mathbf{Y}_{c,i,t} &\le \mathbf{b}_{c,ij}, && \forall j \in \mathcal{N}(i), \\
\mathbf{Y}_{c,\min} &\le \mathbf{Y}_{c,t} \le \mathbf{Y}_{c,\max}, && \forall t \in [0, T],
\end{align}
where $\mathbf{X}_t = [x_t, y_t, v_t, \psi_t]^\top$ denotes the ego state composed of position, velocity, and heading angle, $\mathbf{U}_t = [a_t, \delta_t]^\top$ represents acceleration and steering control, $f_d(\cdot)$ is the discrete kinematic vehicle model, and $\mathcal{N}(i)$ is the set of neighboring agents. Matrices $Q$ and $R$ are positive-definite weighting matrices that regulate trajectory smoothness and control effort. As depicted in Fig.~\ref{fig:planning}, the local QP update iteratively refines the feasible trajectory until convergence to a success solution.

\textbf{Execution and replanning.}
After optimization convergence, only the first segment of the planned trajectory is executed by the low-level controller, while the remainder serves as a predictive reference. At every replanning cycle, new multimodal predictions and environment measurements are incorporated, and the optimization is reinitialized with the previous trajectory as a warm start. This ensures real-time adaptability and stable closed-loop behavior under multimodal uncertainties.
By unifying probabilistic predictions, geometric safety reasoning, and optimization-based refinement, this single-vehicle planning process enables CogDrive to generate trajectories that are dynamically feasible, spatially safe, and cognitively consistent with multimodal interactions.

%
\section{Experiments and Comparative Analysis}
\label{sec:experiments}

\subsection{Experimental Setup}
\label{sec:setup}

\textbf{Datasets.} The proposed CogDrive framework is evaluated on two large-scale, real-world benchmarks: Argoverse 2~\citep{wilson2023argoverse} and INTERACTION~\citep{zhan2019interaction}. Both datasets contain detailed trajectories and high-definition (HD) maps that enable fine-grained evaluation of motion prediction and multi-agent reasoning. The INTERACTION dataset provides naturalistic multi-agent driving data collected in China, Germany, and the United States using drone and roadside sensors. It covers a wide range of complex scenarios including highway ramps, urban intersections, and roundabouts. Each scenario provides centimeter-level lanelet2 maps with lane topology, traffic rule semantics, and connectivity. The dataset includes 18 representative scenarios, each containing diverse vehicle, cyclist, and pedestrian interactions. Every sample contains 1~s of observed trajectory and a 3~s prediction horizon, resulting in approximately 40,000 annotated motion sequences for evaluation.
Argoverse 2 dataset. Argoverse 2 consists of over 250,000 vehicle-centric trajectories collected from various U.S. cities, captured at 10~Hz with accurate localization and map alignment. Compared to its predecessor, it features longer sequences (5~s observation, 6~s prediction) and richer multimodal driving behaviors. The HD maps provide detailed lane geometry, drivable areas, and intersection semantics, making it suitable for evaluating both long-term prediction fidelity and multimodal consistency across urban and highway settings.

\textbf{Evaluation Metrics.} Following standard motion forecasting benchmarks, we adopt both single-agent and joint multi-agent metrics. For Argoverse 2, four metrics are used: minimum Average Displacement Error (minADE), minimum Final Displacement Error (minFDE), Miss Rate (MR), and Brier-minFDE (b-minFDE). For INTERACTION, we report minimum joint metrics, including minimum joint Average Displacement Error (minJointADE) and minimum joint Final Displacement Error (minJointFDE).
Specifically, $\text{minADE}$ computes the average $\ell_2$ distance between predicted and ground-truth positions, while $\text{minFDE}$ focuses on the terminal displacement error. $\text{MR}$ measures the proportion of predictions with a final error exceeding 2.0~m, and $\text{b-minFDE}$ integrates confidence weighting to reflect both accuracy and reliability. The number of predicted modes is fixed at $K=6$ in all experiments.

\subsection{Results and Comparative Analysis}
\label{sec:results}

Tables~\ref{tab:interaction_results} and~\ref{tab:argoverse_results} present quantitative results on the INTERACTION and Argoverse~2 datasets, respectively. CogDrive achieves competitive or superior performance across all key indicators compared with the state-of-the-art methods.

\textbf{Results on INTERACTION.}  
As summarized in Table~\ref{tab:interaction_results}, CogDrive achieves the best minimum joint Final Displacement Error (minJointFDE=0.914~m) and a strong minimum joint Average Displacement Error (minJointADE=0.301~m), surpassing most baselines such as FJMP and HDGT.  
Compared with learning-based models like Trai-MAE and HDGT, CogDrive demonstrates higher consistency in dense multi-agent scenes, indicating its stronger capability to capture complex social interactions and avoid long-tail mispredictions.  
The improvement originates from its cognition-driven multimodal reasoning, which explicitly models behavioral intentions and adapts planning responses to diverse interaction patterns.  
Overall, CogDrive maintains accurate spatial alignment and interpretable trajectory diversity across heterogeneous and highly interactive driving conditions.

\begin{table}[!t]
\centering
\small 
\caption{Trajectory prediction results on the INTERACTION dataset.}
\label{tab:interaction_results}
\begin{tabular}{lcc}
\toprule
\textbf{Method} & \textbf{minJointFDE (m)$\downarrow$} & \textbf{minJointADE (m)$\downarrow$} \\
\midrule
AutoBot~\citep{girgis2021latent} & 1.015 & 0.312 \\
THOMAS~\citep{gilles2021thomas} & 0.968 & 0.416 \\
Trai-MAE~\citep{chen2023traj} & 0.966 & 0.307 \\
HDGT~\citep{jia2023hdgt} & 0.958 & 0.303 \\
FJMP~\citep{rowe2023fjmp} & 0.922 & \textbf{0.275} \\
\textbf{CogDrive (ours)} & \textbf{0.914} & 0.301 \\
\bottomrule
\end{tabular}
\end{table}

\textbf{Results on Argoverse 2.}  
Table~\ref{tab:argoverse_results} presents the quantitative comparisons on the Argoverse 2 dataset. CogDrive achieves the lowest Miss Rate (MR=\textbf{0.120}), indicating the highest safety consistency, while also attaining the best displacement accuracy (b-minFDE=\textbf{1.833~m}, minFDE=\textbf{1.209~m}) and a competitive minADE of 0.803~m.  
Compared with DenseTNT and SceneTransformer, CogDrive yields a lower error in long-horizon trajectories, reflecting enhanced temporal stability and robustness under multimodal uncertainty.  
These improvements benefit from the unified prediction–planning coupling, where multimodal intent reasoning enables the model to anticipate interactions and refine feasible motion trajectories adaptively.  
Such cognition-driven integration ensures smooth, reliable, and human-like motion behaviors in complex urban environments.

\begin{table}[!t]
\centering
\small 
\setlength{\tabcolsep}{6pt} 
\caption{Trajectory prediction results on the Argoverse 2 dataset.}
\label{tab:argoverse_results}
\begin{tabularx}{\linewidth}{lXXXX}
\toprule
\textbf{Method} & \textbf{b-minFDE (m)$\downarrow$} & \textbf{minFDE (m)$\downarrow$} & \textbf{MR$\downarrow$} & \textbf{minADE (m)$\downarrow$} \\
\midrule
LaneGCN~\citep{liang2020learning} & 2.054 & 1.362 & 0.162 & 0.870 \\
mmTransformer~\citep{liu2021multimodal} & 2.033 & 1.338 & 0.154 & 0.844 \\
DenseTNT~\citep{gu2021densetnt} & 1.976 & 1.282 & 0.126 & 0.882 \\
TPCN~\citep{ye2021tpcn} & 1.929 & 1.244 & 0.133 & 0.815 \\
SceneTransformer~\citep{ngiam2021scene} & 1.887 & 1.232 & 0.126 & \textbf{0.803} \\
\textbf{CogDrive (ours)} & \textbf{1.833} & \textbf{1.209} & \textbf{0.120} &  \textbf{0.803} \\
\bottomrule
\end{tabularx}
\end{table}

\begin{figure}[!t]
\centering
\includegraphics[width=\linewidth]{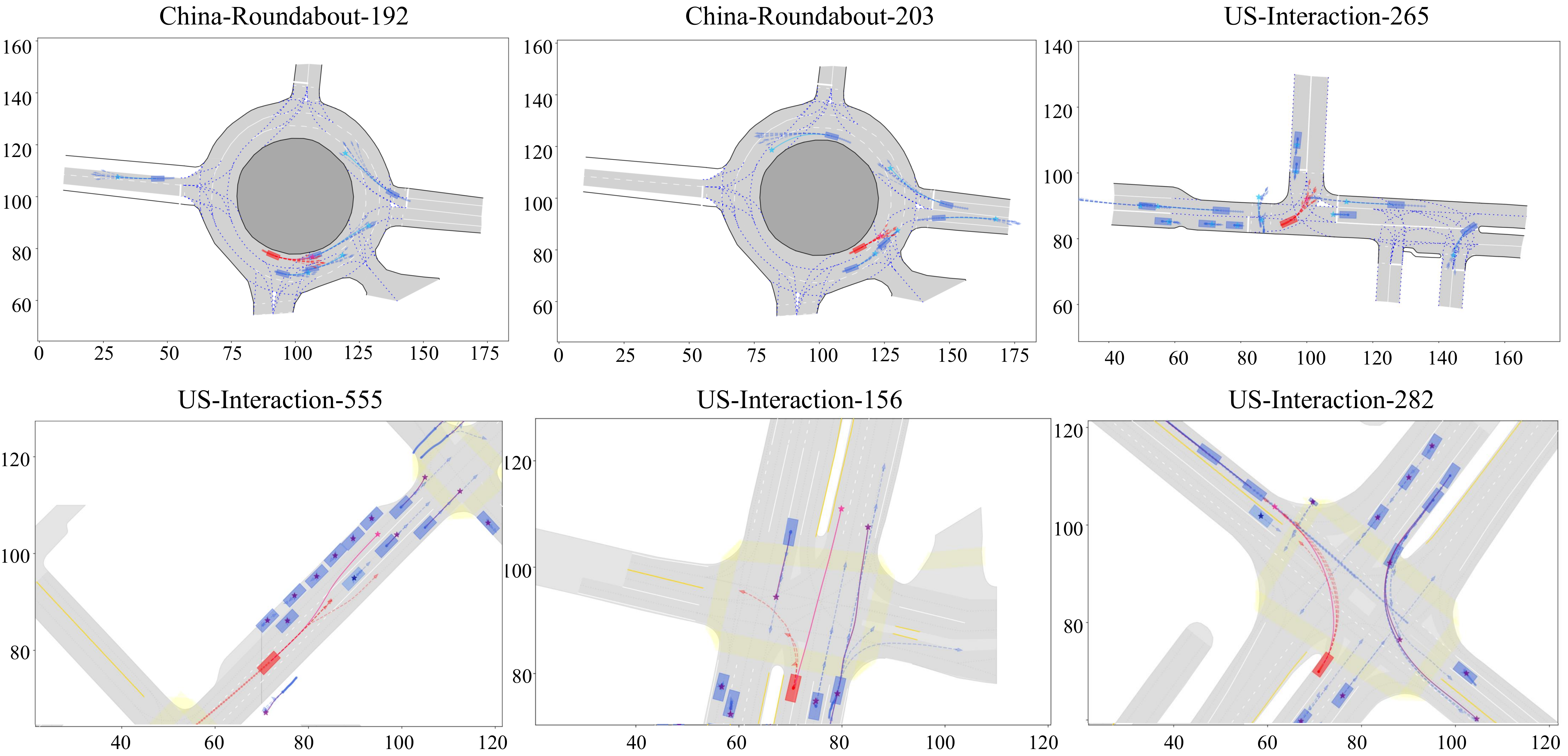}
\caption{Representative multimodal trajectory prediction across diverse driving scenarios. 
The red vehicle represents the ego agent, and colored lines indicate predicted trajectories under different interaction modes. 
The examples span intersections and roundabouts with varying traffic densities and driving behaviors. 
Dots indicate past motion, while solid and dashed lines show multimodal futures. 
CogDrive differentiates behavioral modes, preserves trajectory smoothness, and maintains consistent prediction quality across diverse environments.}
\label{fig:qualitative_multi}
\end{figure}

\textbf{Qualitative analysis.} Figure~\ref{fig:qualitative_multi} presents representative multimodal prediction examples across diverse driving scenarios. The red and blue vehicles denote human-driven agents, with the red vehicle as the ego agent exhibiting multiple potential behaviors. These examples cover various intersection and roundabout configurations under different traffic densities and driver tendencies, where vehicles often engage in yielding, merging, or overtaking interactions. Historical trajectories (dots) represent past motion, while predicted ones (solid and dashed lines) indicate multimodal futures inferred by CogDrive. The ego vehicle may continue, exit, or overtake depending on surrounding dynamics, and CogDrive effectively distinguishes these behavior modes while maintaining spatial smoothness and probabilistic consistency.  

Unlike purely kinematic models, the cognition-driven framework captures both cooperative and competitive interactions, adapting to temporal context and preventing implausible transitions. These results illustrate CogDrive’s ability to generalize across heterogeneous traffic environments and diverse driving cultures, enabling interpretable and safety-consistent multimodal reasoning. 
Overall, CogDrive demonstrates strong generalization and interpretability in representing human-like behavioral uncertainty across heterogeneous, multimodal traffic environments.

\begin{figure}[!t]
\centering
\includegraphics[width=\linewidth]{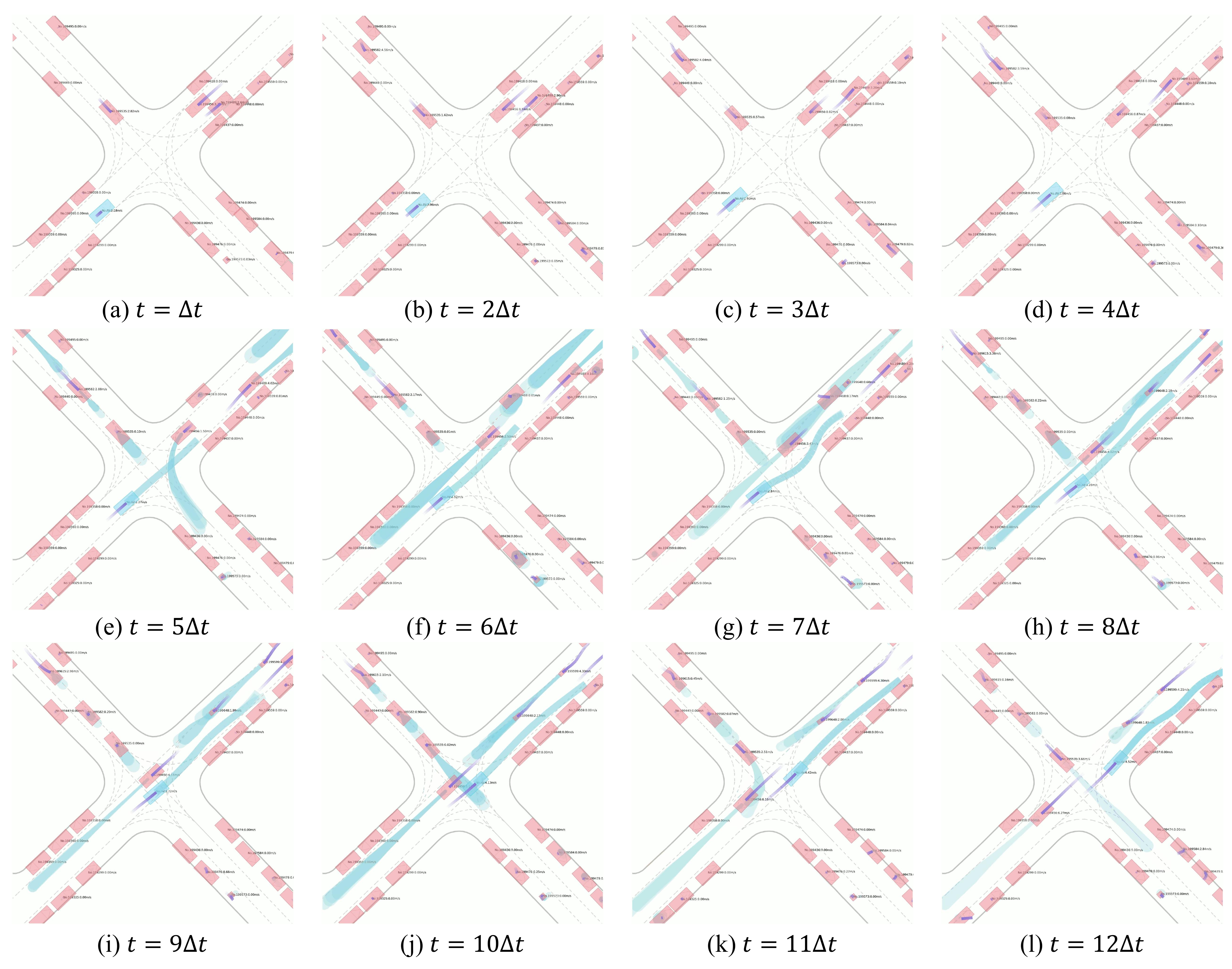}
\caption{Time-sequenced visualization of interactive decision-making at an urban intersection. 
Blue denotes the ego vehicle and red represents surrounding agents. 
Solid lines show observed trajectories, and dashed lines indicate multimodal predictions, where darker tones correspond to higher probability. 
CogDrive performs progressive deceleration, updates mode probabilities, and re-plans short-horizon trajectories to ensure safe and smooth crossing under uncertainty.}
\label{fig:interaction_case}
\end{figure}

\textbf{Interactive Decision-Making Case Study.} To evaluate the decision-making capability of CogDrive, we conduct closed-loop simulations using naturalistic driving data within the MIND framework \citep{li2024multi}, which reproduces real-world geometry and traffic flow consistent with the Argoverse map specifications. In these simulations, surrounding agents follow their recorded trajectories, while the ego vehicle executes online prediction and planning based on CogDrive. This setup enables realistic evaluation of adaptive safety control under multimodal traffic conditions. We focus on intersection scenarios where potential vehicle conflicts occur. Surrounding agents replay their ground-truth motions, and the ego vehicle continuously updates its trajectory as new observations arrive, demonstrating responsive and stable interaction behavior.

Figure~\ref{fig:interaction_case} shows a time-sequenced example at an urban intersection. Blue indicates the ego vehicle, and red denotes surrounding agents. Solid lines represent ground-truth trajectories, while dashed lines correspond to multimodal predictions, with darker tones indicating higher probability. At the start ($t \leq 3\Delta t$), the ego moves at 3.2~m/s and the oncoming vehicle B at 2.8~m/s. Based on early cues, CogDrive infers several hypotheses: whether B will maintain speed, yield, or accelerate. The ego slightly decelerates (0.9~m/s$^2$) to preserve safety margins while refining its belief over modes.  
As the interaction evolves ($t \geq 7\Delta t$), the ego’s confidence shifts toward the yielding behavior of vehicle B. CogDrive triggers its trajectory tree planner, generating short-term alternatives to ensure collision-free passage. It then selects the optimal branch, performing a smooth acceleration (0.7~m/s$^2$) to cross the intersection safely. This case demonstrates how cognition-driven multimodal prediction and planning jointly enable proactive and human-like negotiation under uncertainty. CogDrive anticipates multiple behavioral possibilities, adapts its control in real time, and achieves stable, interpretable decision-making in complex urban environments.

\section{Conclusion}
\label{sec:conclusion}

This paper presents CogDrive, a cognition-driven multimodal prediction and planning fusion framework for safe and adaptive autonomy in complex mixed-traffic environments. By introducing cognitive reasoning principles into motion forecasting and decision-making, CogDrive bridges the gap between data-driven adaptability and rule-based reliability. The framework learns not only from observed motion data but also from cognitive structures that encode how agents interpret and respond to uncertainty. This cognition-driven formulation enhances both behavioral interpretability and generalization under unseen or rare conditions.
The multimodal prediction module captures interaction modes through cognitively inspired representations, linking behavioral semantics with topological motion relationships among agents. Combined with learnable query decoding and differentiable modal learning, it models sparse yet safety-critical behaviors with higher fidelity and robustness. The planning module adopts an emergency-response mechanism and performs safety-stabilized trajectory tree optimization, ensuring that short-term safety and long-term smoothness are maintained across multimodal outcomes. This bidirectional integration transforms prediction and planning into a coherent cognitive process that unifies reasoning, anticipation, and control.
Experiments on the Argoverse 2 and INTERACTION datasets demonstrate that CogDrive achieves strong performance in both accuracy and stability, reducing displacement errors and miss rates while maintaining multimodal consistency. Closed-loop simulations further confirm that the framework produces safe, interpretable, and human-like driving behaviors across diverse interaction scenarios.
Future work will extend CogDrive toward large-scale real-world deployment and human-centered cooperative driving. We envision cognition-aligned autonomy where vehicles reason, anticipate, and adapt like humans to ensure trustworthy and interpretable safety.

\section*{CRediT authorship contribution statement}

Heye Huang: Conceptualization, Formal analysis, Writing – original draft, Visualization. 
Yibin Yang: Software, Writing – review \& editing, Visualization, Validation. Mingfeng Fan: Investigation, Visualization. Haoran Wang: Formal analysis, Writing – review \& editing.  
Xiaocong Zhao: Data curation, Investigation, Writing – review \& editing. 
Jianqiang Wang: Supervision, Conceptualization, Writing – review \& editing.

\section*{Declaration of competing interest}

The authors declare that they have no known competing financial interests or personal relationships that could have appeared to influence the work reported in this paper.

\section*{Acknowledgment}
\label{ack}

We gratefully acknowledge the THICV research group at the School of Vehicle and Mobility, Tsinghua University, for providing the experimental platform, computational resources, and support from all participants involved in the on-road vehicle experiments.

\section*{Data availability}

Data will be made available on request.

\bibliography{reference}

\begin{thebibliography}{46}
\expandafter\ifx\csname natexlab\endcsname\relax\def\natexlab#1{#1}\fi
\providecommand{\url}[1]{\texttt{#1}}
\providecommand{\href}[2]{#2}
\providecommand{\path}[1]{#1}
\providecommand{\DOIprefix}{doi:}
\providecommand{\ArXivprefix}{arXiv:}
\providecommand{\URLprefix}{URL: }
\providecommand{\Pubmedprefix}{pmid:}
\providecommand{\doi}[1]{\href{http://dx.doi.org/#1}{\path{#1}}}
\providecommand{\Pubmed}[1]{\href{pmid:#1}{\path{#1}}}
\providecommand{\bibinfo}[2]{#2}
\ifx\xfnm\relax \def\xfnm[#1]{\unskip,\space#1}\fi
\bibitem[{Cai et~al.(2025)Cai, Lu, Wang, Lian, Chen \& Liu}]{cai2025mlig}
\bibinfo{author}{Cai, Y.}, \bibinfo{author}{Lu, Z.}, \bibinfo{author}{Wang, H.}, \bibinfo{author}{Lian, Y.}, \bibinfo{author}{Chen, L.}, \& \bibinfo{author}{Liu, Q.} (\bibinfo{year}{2025}).
\newblock \bibinfo{title}{Mlig: Scene-level multimodal motion prediction based on multi-layer interaction graph}.
\newblock {\it \bibinfo{journal}{IEEE Transactions on Intelligent Transportation Systems}\/}, .
\bibitem[{Carion et~al.(2020)Carion, Massa, Synnaeve, Usunier, Kirillov \& Zagoruyko}]{carion2020end}
\bibinfo{author}{Carion, N.}, \bibinfo{author}{Massa, F.}, \bibinfo{author}{Synnaeve, G.}, \bibinfo{author}{Usunier, N.}, \bibinfo{author}{Kirillov, A.}, \& \bibinfo{author}{Zagoruyko, S.} (\bibinfo{year}{2020}).
\newblock \bibinfo{title}{End-to-end object detection with transformers}.
\newblock In {\it \bibinfo{booktitle}{European conference on computer vision}\/} (pp. \bibinfo{pages}{213--229}).
\newblock \bibinfo{organization}{Springer}.
\bibitem[{Chen et~al.(2023)Chen, Wang, Shao, Liu, Hao, Guan, Chen \& Heng}]{chen2023traj}
\bibinfo{author}{Chen, H.}, \bibinfo{author}{Wang, J.}, \bibinfo{author}{Shao, K.}, \bibinfo{author}{Liu, F.}, \bibinfo{author}{Hao, J.}, \bibinfo{author}{Guan, C.}, \bibinfo{author}{Chen, G.}, \& \bibinfo{author}{Heng, P.-A.} (\bibinfo{year}{2023}).
\newblock \bibinfo{title}{Traj-mae: Masked autoencoders for trajectory prediction}.
\newblock In {\it \bibinfo{booktitle}{Proceedings of the IEEE/CVF International Conference on Computer Vision}\/} (pp. \bibinfo{pages}{8351--8362}).
\bibitem[{Cheng et~al.(2024)Cheng, Chen, Mei, Yang, Li \& Liu}]{cheng2024rethinking}
\bibinfo{author}{Cheng, J.}, \bibinfo{author}{Chen, Y.}, \bibinfo{author}{Mei, X.}, \bibinfo{author}{Yang, B.}, \bibinfo{author}{Li, B.}, \& \bibinfo{author}{Liu, M.} (\bibinfo{year}{2024}).
\newblock \bibinfo{title}{Rethinking imitation-based planners for autonomous driving}.
\newblock In {\it \bibinfo{booktitle}{2024 IEEE International Conference on Robotics and Automation (ICRA)}\/} (pp. \bibinfo{pages}{14123--14130}).
\newblock \bibinfo{organization}{IEEE}.
\bibitem[{Chib \& Singh(2023)}]{chib2023recent}
\bibinfo{author}{Chib, P.~S.}, \& \bibinfo{author}{Singh, P.} (\bibinfo{year}{2023}).
\newblock \bibinfo{title}{Recent advancements in end-to-end autonomous driving using deep learning: A survey}.
\newblock {\it \bibinfo{journal}{IEEE Transactions on Intelligent Vehicles}\/},  {\it \bibinfo{volume}{9}\/}, \bibinfo{pages}{103--118}.
\bibitem[{Ding et~al.(2021)Ding, Zhang, Chen \& Shen}]{ding2021epsilon}
\bibinfo{author}{Ding, W.}, \bibinfo{author}{Zhang, L.}, \bibinfo{author}{Chen, J.}, \& \bibinfo{author}{Shen, S.} (\bibinfo{year}{2021}).
\newblock \bibinfo{title}{Epsilon: An efficient planning system for automated vehicles in highly interactive environments}.
\newblock {\it \bibinfo{journal}{IEEE Transactions on Robotics}\/},  {\it \bibinfo{volume}{38}\/}, \bibinfo{pages}{1118--1138}.
\bibitem[{Eirale et~al.(2025)Eirale, Leonetti \& Chiaberge}]{eirale2025learning}
\bibinfo{author}{Eirale, A.}, \bibinfo{author}{Leonetti, M.}, \& \bibinfo{author}{Chiaberge, M.} (\bibinfo{year}{2025}).
\newblock \bibinfo{title}{Learning social heuristics for human-aware path planning}.
\newblock {\it \bibinfo{journal}{arXiv preprint arXiv:2509.02134}\/}, .
\bibitem[{Gao et~al.(2020)Gao, Sun, Zhao, Shen, Anguelov, Li \& Schmid}]{gao2020vectornet}
\bibinfo{author}{Gao, J.}, \bibinfo{author}{Sun, C.}, \bibinfo{author}{Zhao, H.}, \bibinfo{author}{Shen, Y.}, \bibinfo{author}{Anguelov, D.}, \bibinfo{author}{Li, C.}, \& \bibinfo{author}{Schmid, C.} (\bibinfo{year}{2020}).
\newblock \bibinfo{title}{Vectornet: Encoding hd maps and agent dynamics from vectorized representation}.
\newblock In {\it \bibinfo{booktitle}{Proceedings of the IEEE/CVF conference on computer vision and pattern recognition}\/} (pp. \bibinfo{pages}{11525--11533}).
\bibitem[{Gilles et~al.(2021)Gilles, Sabatini, Tsishkou, Stanciulescu \& Moutarde}]{gilles2021thomas}
\bibinfo{author}{Gilles, T.}, \bibinfo{author}{Sabatini, S.}, \bibinfo{author}{Tsishkou, D.}, \bibinfo{author}{Stanciulescu, B.}, \& \bibinfo{author}{Moutarde, F.} (\bibinfo{year}{2021}).
\newblock \bibinfo{title}{Thomas: Trajectory heatmap output with learned multi-agent sampling}.
\newblock {\it \bibinfo{journal}{arXiv preprint arXiv:2110.06607}\/}, .
\bibitem[{Girgis et~al.(2021)Girgis, Golemo, Codevilla, Weiss, D'Souza, Kahou, Heide \& Pal}]{girgis2021latent}
\bibinfo{author}{Girgis, R.}, \bibinfo{author}{Golemo, F.}, \bibinfo{author}{Codevilla, F.}, \bibinfo{author}{Weiss, M.}, \bibinfo{author}{D'Souza, J.~A.}, \bibinfo{author}{Kahou, S.~E.}, \bibinfo{author}{Heide, F.}, \& \bibinfo{author}{Pal, C.} (\bibinfo{year}{2021}).
\newblock \bibinfo{title}{Latent variable sequential set transformers for joint multi-agent motion prediction}.
\newblock {\it \bibinfo{journal}{arXiv preprint arXiv:2104.00563}\/}, .
\bibitem[{Gu et~al.(2021)Gu, Sun \& Zhao}]{gu2021densetnt}
\bibinfo{author}{Gu, J.}, \bibinfo{author}{Sun, C.}, \& \bibinfo{author}{Zhao, H.} (\bibinfo{year}{2021}).
\newblock \bibinfo{title}{Densetnt: End-to-end trajectory prediction from dense goal sets}.
\newblock In {\it \bibinfo{booktitle}{Proceedings of the IEEE/CVF international conference on computer vision}\/} (pp. \bibinfo{pages}{15303--15312}).
\bibitem[{Gupta et~al.(2018)Gupta, Johnson, Fei-Fei, Savarese \& Alahi}]{gupta2018social}
\bibinfo{author}{Gupta, A.}, \bibinfo{author}{Johnson, J.}, \bibinfo{author}{Fei-Fei, L.}, \bibinfo{author}{Savarese, S.}, \& \bibinfo{author}{Alahi, A.} (\bibinfo{year}{2018}).
\newblock \bibinfo{title}{Social gan: Socially acceptable trajectories with generative adversarial networks}.
\newblock In {\it \bibinfo{booktitle}{Proceedings of the IEEE conference on computer vision and pattern recognition}\/} (pp. \bibinfo{pages}{2255--2264}).
\bibitem[{Huang et~al.(2025{\natexlab{a}})Huang, Cheng, Zhou, Wang, Liu \& Li}]{huang2025react}
\bibinfo{author}{Huang, H.}, \bibinfo{author}{Cheng, H.}, \bibinfo{author}{Zhou, Z.}, \bibinfo{author}{Wang, Z.}, \bibinfo{author}{Liu, Q.}, \& \bibinfo{author}{Li, X.} (\bibinfo{year}{2025}{\natexlab{a}}).
\newblock \bibinfo{title}{React: Runtime-enabled active collision-avoidance technique for autonomous driving}.
\newblock {\it \bibinfo{journal}{arXiv preprint arXiv:2505.11474}\/}, .
\bibitem[{Huang et~al.(2025{\natexlab{b}})Huang, Liu, Zhang, Zhao, Li \& Wang}]{huang2025lead}
\bibinfo{author}{Huang, H.}, \bibinfo{author}{Liu, J.}, \bibinfo{author}{Zhang, B.}, \bibinfo{author}{Zhao, S.}, \bibinfo{author}{Li, B.}, \& \bibinfo{author}{Wang, J.} (\bibinfo{year}{2025}{\natexlab{b}}).
\newblock \bibinfo{title}{Lead: Learning-enhanced adaptive decision-making for autonomous driving in dynamic environments}.
\newblock {\it \bibinfo{journal}{IEEE Transactions on Intelligent Transportation Systems}\/}, .
\bibitem[{Huang et~al.(2024)Huang, Liu, Liu, Yang, Wang, Abbink \& Zgonnikov}]{huang2024general}
\bibinfo{author}{Huang, H.}, \bibinfo{author}{Liu, Y.}, \bibinfo{author}{Liu, J.}, \bibinfo{author}{Yang, Q.}, \bibinfo{author}{Wang, J.}, \bibinfo{author}{Abbink, D.}, \& \bibinfo{author}{Zgonnikov, A.} (\bibinfo{year}{2024}).
\newblock \bibinfo{title}{General optimal trajectory planning: enabling autonomous vehicles with the principle of least action}.
\newblock {\it \bibinfo{journal}{Engineering}\/},  {\it \bibinfo{volume}{33}\/}, \bibinfo{pages}{63--76}.
\bibitem[{Huang et~al.(2020)Huang, Wang, Fei, Zheng, Yang, Liu, Wu \& Xu}]{huang2020probabilistic}
\bibinfo{author}{Huang, H.}, \bibinfo{author}{Wang, J.}, \bibinfo{author}{Fei, C.}, \bibinfo{author}{Zheng, X.}, \bibinfo{author}{Yang, Y.}, \bibinfo{author}{Liu, J.}, \bibinfo{author}{Wu, X.}, \& \bibinfo{author}{Xu, Q.} (\bibinfo{year}{2020}).
\newblock \bibinfo{title}{A probabilistic risk assessment framework considering lane-changing behavior interaction}.
\newblock {\it \bibinfo{journal}{Science China Information Sciences}\/},  {\it \bibinfo{volume}{63}\/}, \bibinfo{pages}{190203}.
\bibitem[{Hwang et~al.(2024)Hwang, Xu, Lin, Hung, Ji, Choi, Huang, He, Covington, Sapp et~al.}]{hwang2024emma}
\bibinfo{author}{Hwang, J.-J.}, \bibinfo{author}{Xu, R.}, \bibinfo{author}{Lin, H.}, \bibinfo{author}{Hung, W.-C.}, \bibinfo{author}{Ji, J.}, \bibinfo{author}{Choi, K.}, \bibinfo{author}{Huang, D.}, \bibinfo{author}{He, T.}, \bibinfo{author}{Covington, P.}, \bibinfo{author}{Sapp, B.} et~al. (\bibinfo{year}{2024}).
\newblock \bibinfo{title}{Emma: End-to-end multimodal model for autonomous driving}.
\newblock {\it \bibinfo{journal}{arXiv preprint arXiv:2410.23262}\/}, .
\bibitem[{Jia et~al.(2023)Jia, Wu, Chen, Liu, Li \& Yan}]{jia2023hdgt}
\bibinfo{author}{Jia, X.}, \bibinfo{author}{Wu, P.}, \bibinfo{author}{Chen, L.}, \bibinfo{author}{Liu, Y.}, \bibinfo{author}{Li, H.}, \& \bibinfo{author}{Yan, J.} (\bibinfo{year}{2023}).
\newblock \bibinfo{title}{Hdgt: Heterogeneous driving graph transformer for multi-agent trajectory prediction via scene encoding}.
\newblock {\it \bibinfo{journal}{IEEE transactions on pattern analysis and machine intelligence}\/},  {\it \bibinfo{volume}{45}\/}, \bibinfo{pages}{13860--13875}.
\bibitem[{Jiang et~al.(2023)Jiang, Cornman, Park, Sapp, Zhou, Anguelov et~al.}]{jiang2023motiondiffuser}
\bibinfo{author}{Jiang, C.}, \bibinfo{author}{Cornman, A.}, \bibinfo{author}{Park, C.}, \bibinfo{author}{Sapp, B.}, \bibinfo{author}{Zhou, Y.}, \bibinfo{author}{Anguelov, D.} et~al. (\bibinfo{year}{2023}).
\newblock \bibinfo{title}{Motiondiffuser: Controllable multi-agent motion prediction using diffusion}.
\newblock In {\it \bibinfo{booktitle}{Proceedings of the IEEE/CVF conference on computer vision and pattern recognition}\/} (pp. \bibinfo{pages}{9644--9653}).
\bibitem[{Jiang et~al.(2025)Jiang, Zhao, Hu, Chen \& Zhang}]{jiang2025multi}
\bibinfo{author}{Jiang, H.}, \bibinfo{author}{Zhao, B.}, \bibinfo{author}{Hu, C.}, \bibinfo{author}{Chen, H.}, \& \bibinfo{author}{Zhang, X.} (\bibinfo{year}{2025}).
\newblock \bibinfo{title}{Multi-modal vehicle motion prediction based on motion-query social transformer network for internet of vehicles}.
\newblock {\it \bibinfo{journal}{IEEE Internet of Things Journal}\/}, .
\bibitem[{Lenz et~al.(2016)Lenz, Kessler \& Knoll}]{lenz2016tactical}
\bibinfo{author}{Lenz, D.}, \bibinfo{author}{Kessler, T.}, \& \bibinfo{author}{Knoll, A.} (\bibinfo{year}{2016}).
\newblock \bibinfo{title}{Tactical cooperative planning for autonomous highway driving using monte-carlo tree search}.
\newblock In {\it \bibinfo{booktitle}{2016 IEEE Intelligent Vehicles Symposium (IV)}\/} (pp. \bibinfo{pages}{447--453}).
\newblock \bibinfo{organization}{IEEE}.
\bibitem[{Li et~al.(2024)Li, Zhang, Liu \& Shen}]{li2024multi}
\bibinfo{author}{Li, T.}, \bibinfo{author}{Zhang, L.}, \bibinfo{author}{Liu, S.}, \& \bibinfo{author}{Shen, S.} (\bibinfo{year}{2024}).
\newblock \bibinfo{title}{Multi-modal integrated prediction and decision-making with adaptive interaction modality explorations}.
\newblock {\it \bibinfo{journal}{arXiv preprint arXiv:2408.13742}\/}, .
\bibitem[{Liang et~al.(2020)Liang, Yang, Hu, Chen, Liao, Feng \& Urtasun}]{liang2020learning}
\bibinfo{author}{Liang, M.}, \bibinfo{author}{Yang, B.}, \bibinfo{author}{Hu, R.}, \bibinfo{author}{Chen, Y.}, \bibinfo{author}{Liao, R.}, \bibinfo{author}{Feng, S.}, \& \bibinfo{author}{Urtasun, R.} (\bibinfo{year}{2020}).
\newblock \bibinfo{title}{Learning lane graph representations for motion forecasting}.
\newblock In {\it \bibinfo{booktitle}{European Conference on Computer Vision}\/} (pp. \bibinfo{pages}{541--556}).
\newblock \bibinfo{organization}{Springer}.
\bibitem[{Liu et~al.(2025)Liu, Huang, Zhao, Shi, Ahn \& Li}]{liu2025risknet}
\bibinfo{author}{Liu, Q.}, \bibinfo{author}{Huang, H.}, \bibinfo{author}{Zhao, S.}, \bibinfo{author}{Shi, L.}, \bibinfo{author}{Ahn, S.}, \& \bibinfo{author}{Li, X.} (\bibinfo{year}{2025}).
\newblock \bibinfo{title}{Risknet: Interaction-aware risk forecasting for autonomous driving in long-tail scenarios}.
\newblock {\it \bibinfo{journal}{arXiv preprint arXiv:2504.15541}\/}, .
\bibitem[{Liu et~al.(2021)Liu, Zhang, Fang, Jiang \& Zhou}]{liu2021multimodal}
\bibinfo{author}{Liu, Y.}, \bibinfo{author}{Zhang, J.}, \bibinfo{author}{Fang, L.}, \bibinfo{author}{Jiang, Q.}, \& \bibinfo{author}{Zhou, B.} (\bibinfo{year}{2021}).
\newblock \bibinfo{title}{Multimodal motion prediction with stacked transformers}.
\newblock In {\it \bibinfo{booktitle}{Proceedings of the IEEE/CVF conference on computer vision and pattern recognition}\/} (pp. \bibinfo{pages}{7577--7586}).
\bibitem[{Meng et~al.(2021)Meng, Liu, Jing \& Zu}]{meng2021fsm}
\bibinfo{author}{Meng, F.}, \bibinfo{author}{Liu, A.}, \bibinfo{author}{Jing, S.}, \& \bibinfo{author}{Zu, Y.} (\bibinfo{year}{2021}).
\newblock \bibinfo{title}{Fsm trajectory tracking controllers of ob-auv in the horizontal plane}.
\newblock In {\it \bibinfo{booktitle}{2021 IEEE International Conference on Intelligence and Safety for Robotics (ISR)}\/} (pp. \bibinfo{pages}{204--208}).
\newblock \bibinfo{organization}{IEEE}.
\bibitem[{Ngiam et~al.(2021)Ngiam, Caine, Vasudevan, Zhang, Chiang, Ling, Roelofs, Bewley, Liu, Venugopal et~al.}]{ngiam2021scene}
\bibinfo{author}{Ngiam, J.}, \bibinfo{author}{Caine, B.}, \bibinfo{author}{Vasudevan, V.}, \bibinfo{author}{Zhang, Z.}, \bibinfo{author}{Chiang, H.-T.~L.}, \bibinfo{author}{Ling, J.}, \bibinfo{author}{Roelofs, R.}, \bibinfo{author}{Bewley, A.}, \bibinfo{author}{Liu, C.}, \bibinfo{author}{Venugopal, A.} et~al. (\bibinfo{year}{2021}).
\newblock \bibinfo{title}{Scene transformer: A unified architecture for predicting multiple agent trajectories}.
\newblock {\it \bibinfo{journal}{arXiv preprint arXiv:2106.08417}\/}, .
\bibitem[{Ouyang(2024)}]{ouyang2024deyo}
\bibinfo{author}{Ouyang, H.} (\bibinfo{year}{2024}).
\newblock \bibinfo{title}{Deyo: Detr with yolo for end-to-end object detection}.
\newblock {\it \bibinfo{journal}{arXiv preprint arXiv:2402.16370}\/}, .
\bibitem[{Pek \& Althoff(2020)}]{pek2020failsafe}
\bibinfo{author}{Pek, C.}, \& \bibinfo{author}{Althoff, M.} (\bibinfo{year}{2020}).
\newblock \bibinfo{title}{Fail-safe motion planning for online verification of autonomous vehicles using convex optimization}.
\newblock {\it \bibinfo{journal}{IEEE Transactions on Robotics}\/},  {\it \bibinfo{volume}{37}\/}, \bibinfo{pages}{798--814}.
\bibitem[{Prakash et~al.(2021)Prakash, Chitta \& Geiger}]{prakash2021multi}
\bibinfo{author}{Prakash, A.}, \bibinfo{author}{Chitta, K.}, \& \bibinfo{author}{Geiger, A.} (\bibinfo{year}{2021}).
\newblock \bibinfo{title}{Multi-modal fusion transformer for end-to-end autonomous driving}.
\newblock In {\it \bibinfo{booktitle}{Proceedings of the IEEE/CVF conference on computer vision and pattern recognition}\/} (pp. \bibinfo{pages}{7077--7087}).
\bibitem[{Qi et~al.(2017)Qi, Su, Mo \& Guibas}]{qi2017pointnet}
\bibinfo{author}{Qi, C.~R.}, \bibinfo{author}{Su, H.}, \bibinfo{author}{Mo, K.}, \& \bibinfo{author}{Guibas, L.~J.} (\bibinfo{year}{2017}).
\newblock \bibinfo{title}{Pointnet: Deep learning on point sets for 3d classification and segmentation}.
\newblock In {\it \bibinfo{booktitle}{Proceedings of the IEEE conference on computer vision and pattern recognition}\/} (pp. \bibinfo{pages}{652--660}).
\bibitem[{Rowe et~al.(2023)Rowe, Ethier, Dykhne \& Czarnecki}]{rowe2023fjmp}
\bibinfo{author}{Rowe, L.}, \bibinfo{author}{Ethier, M.}, \bibinfo{author}{Dykhne, E.-H.}, \& \bibinfo{author}{Czarnecki, K.} (\bibinfo{year}{2023}).
\newblock \bibinfo{title}{Fjmp: Factorized joint multi-agent motion prediction over learned directed acyclic interaction graphs}.
\newblock In {\it \bibinfo{booktitle}{Proceedings of the IEEE/CVF Conference on Computer Vision and Pattern Recognition}\/} (pp. \bibinfo{pages}{13745--13755}).
\bibitem[{Salzmann et~al.(2020)Salzmann, Ivanovic, Chakravarty \& Pavone}]{salzmann2020trajectron++}
\bibinfo{author}{Salzmann, T.}, \bibinfo{author}{Ivanovic, B.}, \bibinfo{author}{Chakravarty, P.}, \& \bibinfo{author}{Pavone, M.} (\bibinfo{year}{2020}).
\newblock \bibinfo{title}{Trajectron++: Dynamically-feasible trajectory forecasting with heterogeneous data}.
\newblock In {\it \bibinfo{booktitle}{European Conference on Computer Vision}\/} (pp. \bibinfo{pages}{683--700}).
\newblock \bibinfo{organization}{Springer}.
\bibitem[{Shaoul et~al.(2024)Shaoul, Mishani, Vats, Li \& Likhachev}]{shaoul2024multi}
\bibinfo{author}{Shaoul, Y.}, \bibinfo{author}{Mishani, I.}, \bibinfo{author}{Vats, S.}, \bibinfo{author}{Li, J.}, \& \bibinfo{author}{Likhachev, M.} (\bibinfo{year}{2024}).
\newblock \bibinfo{title}{Multi-robot motion planning with diffusion models}.
\newblock {\it \bibinfo{journal}{arXiv preprint arXiv:2410.03072}\/}, .
\bibitem[{Sheng et~al.(2024)Sheng, Yu, Parker, Kwiatkowska \& Feng}]{sheng2024safe}
\bibinfo{author}{Sheng, S.}, \bibinfo{author}{Yu, P.}, \bibinfo{author}{Parker, D.}, \bibinfo{author}{Kwiatkowska, M.}, \& \bibinfo{author}{Feng, L.} (\bibinfo{year}{2024}).
\newblock \bibinfo{title}{Safe pomdp online planning among dynamic agents via adaptive conformal prediction}.
\newblock {\it \bibinfo{journal}{IEEE Robotics and Automation Letters}\/}, .
\bibitem[{Shi et~al.(2022)Shi, Jiang, Dai \& Schiele}]{shi2022motion}
\bibinfo{author}{Shi, S.}, \bibinfo{author}{Jiang, L.}, \bibinfo{author}{Dai, D.}, \& \bibinfo{author}{Schiele, B.} (\bibinfo{year}{2022}).
\newblock \bibinfo{title}{Motion transformer with global intention localization and local movement refinement}.
\newblock {\it \bibinfo{journal}{Advances in Neural Information Processing Systems}\/},  {\it \bibinfo{volume}{35}\/}, \bibinfo{pages}{6531--6543}.
\bibitem[{Sun et~al.(2024)Sun, Yuan, Sun, Wang, Han, Ma, Huang, Wong, Tee \& Ang}]{sun2024controlmtr}
\bibinfo{author}{Sun, J.}, \bibinfo{author}{Yuan, C.}, \bibinfo{author}{Sun, S.}, \bibinfo{author}{Wang, S.}, \bibinfo{author}{Han, Y.}, \bibinfo{author}{Ma, S.}, \bibinfo{author}{Huang, Z.}, \bibinfo{author}{Wong, A.}, \bibinfo{author}{Tee, K.~P.}, \& \bibinfo{author}{Ang, M.~H.} (\bibinfo{year}{2024}).
\newblock \bibinfo{title}{Controlmtr: Control-guided motion transformer with scene-compliant intention points for feasible motion prediction}.
\newblock In {\it \bibinfo{booktitle}{2024 IEEE 27th International Conference on Intelligent Transportation Systems (ITSC)}\/} (pp. \bibinfo{pages}{1507--1514}).
\newblock \bibinfo{organization}{IEEE}.
\bibitem[{Wilson et~al.(2023)Wilson, Qi, Agarwal, Lambert, Singh, Khandelwal, Pan, Kumar, Hartnett, Pontes et~al.}]{wilson2023argoverse}
\bibinfo{author}{Wilson, B.}, \bibinfo{author}{Qi, W.}, \bibinfo{author}{Agarwal, T.}, \bibinfo{author}{Lambert, J.}, \bibinfo{author}{Singh, J.}, \bibinfo{author}{Khandelwal, S.}, \bibinfo{author}{Pan, B.}, \bibinfo{author}{Kumar, R.}, \bibinfo{author}{Hartnett, A.}, \bibinfo{author}{Pontes, J.~K.} et~al. (\bibinfo{year}{2023}).
\newblock \bibinfo{title}{Argoverse 2: Next generation datasets for self-driving perception and forecasting}.
\newblock {\it \bibinfo{journal}{arXiv preprint arXiv:2301.00493}\/}, .
\bibitem[{Yang et~al.(2024)Yang, Xu, Yan, Jiang, Wang \& Huang}]{yang2024csdo}
\bibinfo{author}{Yang, Y.}, \bibinfo{author}{Xu, S.}, \bibinfo{author}{Yan, X.}, \bibinfo{author}{Jiang, J.}, \bibinfo{author}{Wang, J.}, \& \bibinfo{author}{Huang, H.} (\bibinfo{year}{2024}).
\newblock \bibinfo{title}{Csdo: Enhancing efficiency and success in large-scale multi-vehicle trajectory planning}.
\newblock {\it \bibinfo{journal}{IEEE Robotics and Automation Letters}\/}, .
\bibitem[{Ye et~al.(2021)Ye, Cao \& Chen}]{ye2021tpcn}
\bibinfo{author}{Ye, M.}, \bibinfo{author}{Cao, T.}, \& \bibinfo{author}{Chen, Q.} (\bibinfo{year}{2021}).
\newblock \bibinfo{title}{Tpcn: Temporal point cloud networks for motion forecasting}.
\newblock In {\it \bibinfo{booktitle}{Proceedings of the IEEE/CVF Conference on Computer Vision and Pattern Recognition}\/} (pp. \bibinfo{pages}{11318--11327}).
\bibitem[{Zhan et~al.(2019)Zhan, Sun, Wang, Shi, Clausse, Naumann, Kummerle, Konigshof, Stiller, de~La~Fortelle et~al.}]{zhan2019interaction}
\bibinfo{author}{Zhan, W.}, \bibinfo{author}{Sun, L.}, \bibinfo{author}{Wang, D.}, \bibinfo{author}{Shi, H.}, \bibinfo{author}{Clausse, A.}, \bibinfo{author}{Naumann, M.}, \bibinfo{author}{Kummerle, J.}, \bibinfo{author}{Konigshof, H.}, \bibinfo{author}{Stiller, C.}, \bibinfo{author}{de~La~Fortelle, A.} et~al. (\bibinfo{year}{2019}).
\newblock \bibinfo{title}{Interaction dataset: An international, adversarial and cooperative motion dataset in interactive driving scenarios with semantic maps}.
\newblock {\it \bibinfo{journal}{arXiv preprint arXiv:1910.03088}\/}, .
\bibitem[{Zhang et~al.(2024)Zhang, Li, Liu \& Shen}]{zhang2024simpl}
\bibinfo{author}{Zhang, L.}, \bibinfo{author}{Li, P.}, \bibinfo{author}{Liu, S.}, \& \bibinfo{author}{Shen, S.} (\bibinfo{year}{2024}).
\newblock \bibinfo{title}{Simpl: A simple and efficient multi-agent motion prediction baseline for autonomous driving}.
\newblock {\it \bibinfo{journal}{IEEE Robotics and Automation Letters}\/},  {\it \bibinfo{volume}{9}\/}, \bibinfo{pages}{3767--3774}.
\bibitem[{Zhang et~al.(2023)Zhang, Liniger, Sakaridis, Yu \& Gool}]{zhang2023real}
\bibinfo{author}{Zhang, Z.}, \bibinfo{author}{Liniger, A.}, \bibinfo{author}{Sakaridis, C.}, \bibinfo{author}{Yu, F.}, \& \bibinfo{author}{Gool, L.~V.} (\bibinfo{year}{2023}).
\newblock \bibinfo{title}{Real-time motion prediction via heterogeneous polyline transformer with relative pose encoding}.
\newblock {\it \bibinfo{journal}{Advances in Neural Information Processing Systems}\/},  {\it \bibinfo{volume}{36}\/}, \bibinfo{pages}{57481--57499}.
\bibitem[{Zhou et~al.(2023)Zhou, Wang, Li \& Huang}]{zhou2023query}
\bibinfo{author}{Zhou, Z.}, \bibinfo{author}{Wang, J.}, \bibinfo{author}{Li, Y.-H.}, \& \bibinfo{author}{Huang, Y.-K.} (\bibinfo{year}{2023}).
\newblock \bibinfo{title}{Query-centric trajectory prediction}.
\newblock In {\it \bibinfo{booktitle}{Proceedings of the IEEE/CVF conference on computer vision and pattern recognition}\/} (pp. \bibinfo{pages}{17863--17873}).
\bibitem[{Zhou et~al.(2022{\natexlab{a}})Zhou, Ye, Wang, Wu \& Lu}]{zhou2022hivt}
\bibinfo{author}{Zhou, Z.}, \bibinfo{author}{Ye, L.}, \bibinfo{author}{Wang, J.}, \bibinfo{author}{Wu, K.}, \& \bibinfo{author}{Lu, K.} (\bibinfo{year}{2022}{\natexlab{a}}).
\newblock \bibinfo{title}{Hivt: Hierarchical vector transformer for multi-agent motion prediction}.
\newblock In {\it \bibinfo{booktitle}{Proceedings of the IEEE/CVF conference on computer vision and pattern recognition}\/} (pp. \bibinfo{pages}{8823--8833}).
\bibitem[{Zhou et~al.(2022{\natexlab{b}})Zhou, Ye, Wang, Wu \& Lu}]{zhou2022hierarchical}
\bibinfo{author}{Zhou, Z.}, \bibinfo{author}{Ye, L.}, \bibinfo{author}{Wang, J.}, \bibinfo{author}{Wu, K.}, \& \bibinfo{author}{Lu, K.~H.} (\bibinfo{year}{2022}{\natexlab{b}}).
\newblock \bibinfo{title}{Hierarchical vector transformer for multi-agent motion prediction. in 2022 ieee}.
\newblock In {\it \bibinfo{booktitle}{CVF Conference on Computer Vision and Pattern Recognition (CVPR)}\/} (pp. \bibinfo{pages}{8813--8823}).

\end{thebibliography}

\end{document}